\documentclass[10pt,twocolumn,letterpaper]{article}
\DeclareMathAlphabet{\mathbbold}{U}{bbold}{m}{n} 
\usepackage{cvpr}      

\usepackage[accsupp]{axessibility}
\usepackage[T1]{fontenc}
\usepackage{pifont}
\usepackage{graphicx}
\usepackage{glossaries}
\glsdisablehyper
\usepackage{pgfplots}
\usepackage{pgfplotstable}
\usepackage{multirow}
\usepackage{tikz}
\usetikzlibrary{shadows,shadows.blur,patterns}
\usepackage{cuted}
\usepackage{ragged2e}
\usepackage{algorithm}
\usepackage[noend]{algpseudocode}

\usepackage{helvet}
\usepackage[eulergreek]{sansmath}

\tikzfading[name=fade down,
    top color=transparent!0,
    bottom color=transparent!20]

\usepgfplotslibrary{fillbetween,colorbrewer}

\pgfplotsset{compat=1.18} 

\pgfdeclareplotmark{01}{\pgfuseplotmark{+}}
\pgfdeclareplotmark{02}{\pgfuseplotmark{triangle}}
\pgfdeclareplotmark{03}{\pgfuseplotmark{square}}

\pgfplotscreateplotcyclelist{scatter}{%
    {Aquamarine,fill=Aquamarine!100!white,draw=white, draw opacity = 0.4},
    {Orchid,fill=Orchid!100!white,draw=white, draw opacity = 0.4},
    {RoyalBlue,fill=RoyalBlue!100!white,draw=white, draw opacity = 0.4},
    {Cyan,fill=Cyan!100!white,draw=white, draw opacity = 0.4},
    {Melon,fill=Melon!100!white,draw=white, draw opacity = 0.4},
    {Goldenrod,fill=Goldenrod!100!white,draw=white, draw opacity = 0.4},
    {VioletRed,fill=VioletRed!100!white,draw=white, draw opacity = 0.4}
}

\pgfplotsset{
    /pgfplots/bar cycle list/.style={/pgfplots/cycle list={
        {Cyan,fill=Cyan!100!white,mark=none},
        {VioletRed,fill=VioletRed!100!white,mark=none},
        {RoyalBlue,fill=RoyalBlue!100!white,mark=none}
        },
    },
}

\pgfplotsset{
    colormap={confmatrix}{
        rgb255=(255, 255, 255);
        rgb255=(130, 245, 255);
        rgb255=(218, 165, 32);
        rgb255=(253, 188, 180);
        rgb255=(238, 130, 238);
        rgb255=(158, 52, 154);
    },
}

\makeatletter \newcommand{\pgfplotsdrawaxis}{\pgfplots@draw@axis} \makeatother

\pgfplotsset{axis line on top/.style={
  axis line style=transparent,
  ticklabel style=transparent,
  tick style=transparent,
  axis on top=false,
  after end axis/.append code={
    \pgfplotsset{
      axis line style=opaque,
      ticklabel style=opaque,
      tick style=opaque,
      grid=none
    }
    \pgfplotsdrawaxis
  }
  }
}

\pgfplotsset{every tick label/.append style={font=\small}}

\newacronym{fer}{FER}{facial expression recognition}
\newacronym{wacc}{Acc.}{accuracy}
\newacronym{f1}{F1}{F1 score}

\newglossaryentry{voxceleb2}{
name=VoxCeleb2,
description=VoxCeleb2 dataset}

\newglossaryentry{meld}{
name=MELD,
description=MELD dataset}

\newglossaryentry{mosei}{
name=CMU-MOSEI,
description=CMU-MOSEI dataset}

\newglossaryentry{caer}{
name=CAER,
description=CAER dataset}

\newglossaryentry{representation_head}{
name=representation head,
description=the network head that produces the representations for downstream trainings}

\newglossaryentry{projection_head}{
name=projection head,
description=the network head that produces projections for pretraining}

\newglossaryentry{clustering_head}{
name=clustering head,
description=the network head that produces the embeddings used for pertaining k-means clustering}

\definecolor{cvprblue}{rgb}{0.21,0.49,0.74}
\usepackage[pagebackref,colorlinks,citecolor=cvprblue]{hyperref}

\def\paperID{15} 
\def\confName{ABAW workshop at CVPR}
\def\confYear{2024}

\title{Multi-Task Multi-Modal Self-Supervised Learning for Facial Expression Recognition}

\author{Marah Halawa\textsuperscript{1,4}\thanks{equal contribution}\ , Florian Blume\textsuperscript{1,4}\footnotemark[1]\ , Pia Bideau\textsuperscript{2}, Martin Maier\textsuperscript{3,4}\\
Rasha Abdel Rahman\textsuperscript{3,4}, Olaf Hellwich\textsuperscript{1,4}\\
\textsuperscript{1}Technische Universität Berlin, \textsuperscript{2}Univ. Grenoble Alpes, Inria, CNRS, Grenoble INP, LJK, \\ \textsuperscript{3}Humboldt-Universität zu Berlin, 
\textsuperscript{4}Science of Intelligence, Research Cluster of Excellence, Berlin\\
{\tt\small \{marah.halawa, florian.blume, olaf.hellwich\}@tu-berlin.de}\\
{\tt\small \{pia.bideau\}@inria.fr, \tt\small \{martin.maier, rasha.abdel.rahman\}@hu-berlin.de}
}

\begin{document}
\maketitle
\begin{abstract}
Human communication is multi-modal; \eg, face-to-face interaction involves auditory signals (speech) and visual signals (face movements and hand gestures). Hence, it is essential to exploit multiple modalities when designing machine learning-based facial expression recognition systems. In addition, given the ever-growing quantities of video data that capture human facial expressions, such systems should utilize raw unlabeled videos without requiring expensive annotations. Therefore, in this work, we employ a multi-task multi-modal self-supervised learning method for facial expression recognition from in-the-wild video data. Our model combines three self-supervised objective functions: First, a multi-modal contrastive loss, that pulls diverse data modalities of the same video together in the representation space. Second, a multi-modal clustering loss that preserves the semantic structure of input data in the representation space. Finally, a multi-modal data reconstruction loss. We conduct a comprehensive study on this multi-modal multi-task self-supervised learning method on three facial expression recognition benchmarks. To that end, we examine the performance of learning through different combinations of self-supervised tasks on the facial expression recognition downstream task. Our model \textbf{ConCluGen} outperforms several multi-modal self-supervised and fully supervised baselines on the \Gls{mosei} dataset. Our results generally show that multi-modal self-supervision tasks offer large performance gains for challenging tasks such as facial expression recognition, while also reducing the amount of manual annotations required. We release our pre-trained models as well as source code publicly \footnote{https://github.com/tub-cv-group/conclugen}.
\end{abstract}    
\section{Introduction}
\label{sec:intro}
\Gls{fer} is a fundamental task for successful everyday human social interaction, and human-computer interaction \cite{6131215}. Rooted in the context-sensitive and top-down manner of human perception, how we perceive an expression can change with (affective) context and prior knowledge \cite{suessPerceivingEmotionsNeutral2015,baumClearJudgmentsBased2020,feldmanbarrettContextRoutinelyEncoded2010}, and other various other factors \cite{wieserFacesContextReview2012}. The same facial expression can be perceived differently depending on the situation and context \cite{russellReadingEmotionsFaces1997,aviezerInherentlyContextualizedNature2017,eiserbeckDeepfakeSmilesMatter2023}. A recent review from \citeauthor{maierKnowledgeaugmentedFacePerception2022} \cite{maierKnowledgeaugmentedFacePerception2022} highlights that to develop \gls{fer} systems that align with human perception, we should consider contextual cues along with social knowledge. From a human perspective, context is inherently multi-modal, not just what is visually perceptible, as often previously treated in computer vision \cite{leeContextAwareEmotionRecognition2019,zhangPuttingVisualObject2020,kostiContextBasedEmotion2019a}.

Over the last decade, deep learning aproaches have advanced the field of artificial intelligence by utilizing the massive amounts of data generated daily, \eg on the internet. Large quantities of such data are multimodal, such as videos. Even though real-world (in the wild) video data helps train deeper machine-learning models, it also presents multiple challenges. Such data is usually imbalanced, noisy, and, most importantly, unlabeled. Therefore, research in deep learning and computer vision directs attention toward self-supervised learning algorithms, that aim to learn rich data representations without the need for manual labeling processes. Self-supervised learning is a form of unsupervised representation learning where the labels are extracted from the data itself, enabling label-efficient feature learning. Subsequently, the resulting models after the self-supervised learning phase can be used or adapted to downstream tasks, such as facial expression recognition. Given the growing amounts of video data that capture human facial expressions, self-supervision may allow for learning data representations from raw unlabeled video samples.
Nevertheless, as mentioned above, \gls{fer} is a challenging task that requires the integration of multi-modal contexts to align with human-level perception, since people their feelings across different modalities (visually, orally, and in other ways). Thus, not only is it essential to include these multiple modalities in learning algorithmss, but we also need to effectively model the interactions across these modalities to enhance \gls{fer}. In this work, we introduce a multi-task multi-modal self-supervised approach for \gls{fer}. Employing multiple tasks for self-supervision allows for learning more informative data representations, as each task enhances a certain property in the learned features, and integrating these tasks allows for capturing these complementary properties in the resulting embedding space. 

\textbf{Contributions}. The main contributions in this paper are as follows:
\begin{itemize}
  \item To the best of our knowledge we are the first to employ multi-task multi-modal self-supervised methods for \gls{fer}. 
  \item Our multi-modal multi-task self-supervised model \textit{ConCluGen} (see \cref{fig:overview}) outperforms multiple self-supervised and fully supervised models on \gls{fer}.
  \item We provide a comprehensive study on multiple self-supervised learning methods for \gls{fer}. 
  \item We make all self-supervised models (multi-modal and uni-modal) presented in this study publicly available to the research community as baselines for future studies. 
\end{itemize}

\section{Related work}
\label{sec:related_work}

\subsection{Self-Supervised Learning}
Over the last years, self-supervised learning methods advanced the research in different fields \cite{bert,pathak19disagreement,Kim2023.04.28.538691,ChantrySchiappa2022SelfSupervisedLF}, especially in computer vision applications where the amount of unlabeled data is increased rapidly due to social media, \eg YouTube videos, TikTok, and TV-series. The reason is that with self-supervised learning methods, we can learn robust feature representation for the input data without the need for annotations. Such approaches generate the labels for the pretext task automatically from the data itself \cite{Larsson_2017}. Different types of self-supervised methods have been developed: Some are predictive (also called generative) \cite{conf/iclr/GidarisSK18}, \ie anticipate and/or generate some parts of the data from another part of the data. The others are contrastive, which aim to anticipate the relations between data samples \cite{simclr}. Some utilize one modality \cite{10.1007/978-3-319-46466-4_5}, while others draw on multiple modalities \cite{Sun2019VideoBERTAJ}. In this paper, we examine different self-supervised approaches and show that especially for complex computer vision applications, such as \gls{fer}, models can benefit from utilizing multi-modal data in the self-supervised learning techniques.\\

\noindent
\textbf{Instance-Level Contrastive Learning.}
Instance-level contrastive learning has been used in multiple domains to improve learned representations from unlabeled data \cite{simclr,word2vec}. The main goal of such algorithms is to pull representations of similar objects together and push dissimilar objects far away from each other. Despite recent achievements of instance-level contrastive methods in multiple domains, they still suffer from downsides such as class collision \cite{classcollision}. In downstream classification tasks, instance-level contrastive learning methods lose semantic similarity between images from the same class. To mitigate this effect, some research used graph-based methods with weak supervision to pull similar instances closer to each other \cite{classcollision}. Others used domain knowledge to create semantically aware positive and negative sets for contrastive learning \cite{ABC}. Moreover, some research focuses on introducing special augmentation techniques that can directly improve a specific downstream task performance rather than focusing on the generality of the self-supervised pre-trained model \cite{Shu2022RevisitingSC}. Additionally, \citeauthor{Doersch2017MultitaskSV} \cite{Doersch2017MultitaskSV} study multi-task self-supervised learning for visual representation learning. They jointly learn four simple but diverse self-supervised methods (motion segmentation, colorization, Exemplar task, and relative position). 
In this paper, we study different self-supervised methods for \gls{fer}. We also study multi-task self-supervised learning, but in different combinations compared to \cite{Doersch2017MultitaskSV}. We also focus more on multi-modality in multi-task self-supervised learning. \\

\noindent
\textbf{Multi-Modal Self-Supervised Learning.}
Self-supervised learning techniques have also advanced multi-modal learning. In such an environment, the model can draw on different data modalities to learn robust representation. Some research applies contrastive techniques in a multi-modal setting for multiple modalities together \cite{cmc}, or from two modalities as in CLIP \cite{CLIP} that was first employed for text and audio representation learning, or to learn from visual and audio \cite{Owens2018AudioVisualSA}. Moreover, \citeauthor{Taleb_2022_CVPR} \cite{Taleb_2022_CVPR} applied a multi-modal self-supervised approach on genetic modalities, and \citeauthor{Enrico} \cite{Enrico} utilize facial landmarks as an additional modality for \gls{fer} in videos. Additionally, a body of research work \cite{clusteringCont,AsanoP0V20,10.5555/3495724.3496542} supports multi-modal contrastive learning with multi-modal clustering to ensure that semantically similar modalities are represented close to each other in the embedding space. In this paper, we follow the same direction in our ConCluGen and ConClu models (\cref{multi_task}). To the best of our knowledge, we are the first to apply such a technique for \gls{fer}. 
\subsection{Facial Expression Recognition}
\Gls{fer} is a challenging task in computer vision for multiple reasons, such as overlapping with other facial features such as identity \cite{survayFER}, and overlapping between multiple expression labels \cite{MOSEI}. Some research focuses on disentangling expression features from other facial features in the input images \cite{halawa2020learning,Liu2018ExploringDF}. Other research focuses on learning a multi-modal representation for the \gls{fer}. In \cite{9206016}, \citeauthor{9206016} introduced a multi-modal deep learning model for \gls{fer} with a mechanism for the late fusion of multi-modal SSL features. The proposed technique is based on transformers and attention mechanisms. Our model differs from \cite{9206016} in that we learn a joint representation space for all input modalities. In \cite{Enrico}, the authors used a pairwise contrastive objective function inspired by CLIP \cite{CLIP} to learn unsupervised multi-modal representations from video data paired with text, audio, and facial landmarks. Our model is different from \cite{Enrico} in that we preserve the semantic similarity between all modalities by learning a multi-modal clustering objective that regularizes the pairwise multi-modal contrastive objective.

Other works focus on the late fusion techniques to fuse different representations from different input modalities efficiently into one multi-modal representation that is free of redundant information. The latter is less prone to overfitting in the prediction task \cite{CMIB}. To the best of our knowledge, we are the first to study multi-task multi-modal self-supervised learning for \gls{fer}. In this work, we study the combination of multiple self-supervised losses, including multi-modal contrastive learning with clustering, instance-based contrastive learning, and generative self-supervised learning.
\section{Methodology}
\label{sec:methods}

\begin{figure*}
    \centering
    \includegraphics[width=\textwidth]{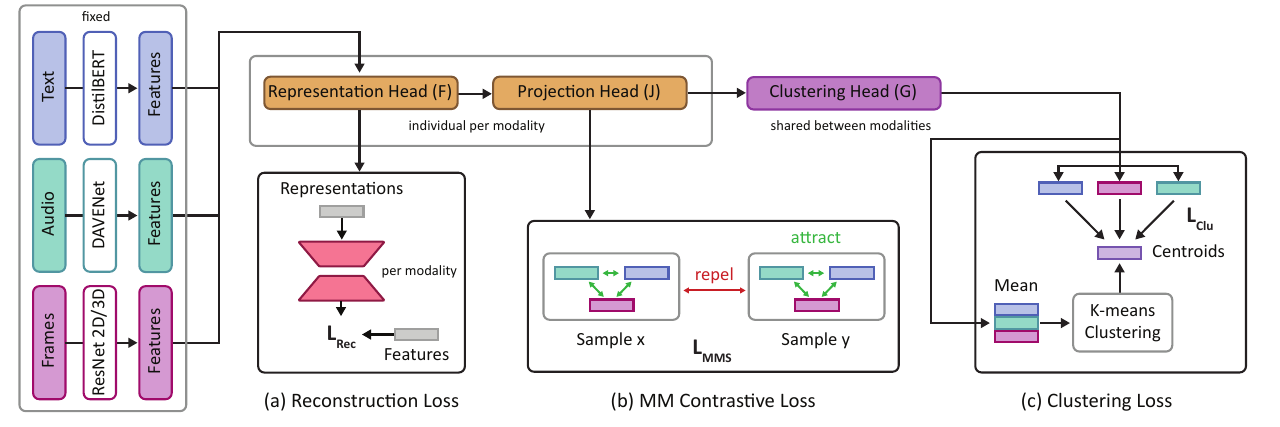}
    \caption{Overview of the architecture of the multi-task multi-modal self-supervised method. The backbone feature extractors process the input modalities in blocks, which we average over the time domain. The resulting temporal features are stored on disk, \ie the backbones are fully fixed. The 3-layer \gls{representation_head} produces the representations we want to use in downstream training. The three losses are: (a) A reconstruction loss which reconstructs the features of each modality individually. (b) The multi-modal contrastive loss encourages the representations from the projection head for modalities belonging to the same input to be represented closer to each other. (c) The multi-modal clustering loss which drives the modalities of a sample towards the centroids computed by k-means clustering. The latter uses the mean of the modalities to compute these centroids.}
    \label{fig:overview}
\end{figure*}

In this section, we will explain the self-supervised pre-training techniques that we adopt in this work. Then we will explain our multi-task multimodal self-supervised model.

\subsection{Problem Formulation}
\label{methods_problem_formulation}

Each input is an utterance that consists of a segment of a video as a sequence of frames $V = \left ( \mathbf{v_1}, \dotsc, \mathbf{v_n} \right )$, a corresponding audio spectrum $A= \left ( \mathbf{a_1}, \dotsc, \mathbf{a_m} \right )$, and a corresponding text subtitle $T = \left ( \mathbf{t_1}, \dotsc, \mathbf{t_k} \right )$. The task is first to learn a rich representation for the input in an unsupervised manner, then predict the facial expression class of the corresponding input video. Due to the unsupervised nature of our approach, we can build a \gls{fer} model without the need for a huge dataset.  We evaluate the following self-supervised methods on both multi-label and single-label \gls{fer} using three datasets. Note that that both pretext and downstream tasks are independent. Each input instance $I$ is represented by a triple input $(\mathbf{v_i},\mathbf{t_i},\mathbf{a_i})$ where $i \in \left \{ 0,N \right \}$ and $N$ is the total number of input samples. We learn a representation for each modality by mapping each input modality to a lower dimensional space using three separate encoders as follows: A video encoder $F_v(\mathbf{v}) \Rightarrow  \mathbf{D_v}$ that maps each video frame sequence to lower dimensional $\mathbf{D_v}$. A text encoder $F_t(\mathbf{t}) \Rightarrow \mathbf{D_t}$ that maps each text subtitle sequence to lower dimensional $\mathbf{D_t}$. Finally, an audio encoder $F_a(\mathbf{a}) \Rightarrow \mathbf{D_a}$ that maps each audio spectrum sequence to lower dimensional $\mathbf{D_a}$. To learn contrastive representations, we map $\mathbf{D_v}$, $\mathbf{D_t}$, and $\mathbf{D_a}$ representations into three separate projection heads (one for each modality), each consisting of two linear layers, as follows: $J_{\mathbf{v}}(\mathbf{D_v}) \Rightarrow  \mathbf{P_v}$,  $J_{\mathbf{t}}(\mathbf{D_t}) \Rightarrow \mathbf{P_t}$, and $J_{\mathbf{a}}(\mathbf{D_a}) \Rightarrow \mathbf{P_a}$. After that, to learn the multi-modal clustering, we map the output of the projection heads to one layer clustering head $G$ (see \cref{fig:overview}), where $G(\mathbf{D_v}) \Rightarrow \mathbf{g_v}$, $G(\mathbf{D_t}) \Rightarrow \mathbf{g_t}$, and $G(\mathbf{D_a}) \Rightarrow \mathbf{g_a}$. \cref{fig:overview} depicts an overview of this architecture. We named the encoders $F$ \textit{Representation Head} here. 

Note that the feature extraction networks in \cref{fig:overview} are fully fixed since we focus on improving the representations obtained from $F$. This means that in the previous paragraphs, $(\mathbf{v_i}, \mathbf{t_i}, \mathbf{a_i})$ are not the raw data, but the fixed output features extracted from 2D and 3D ResNet (\cite{He2015DeepRL} and \cite{pretorchedx}), DistilBERT \cite{Sanh2019DistilBERTAD}, and DAVENet \cite{Harwath2018JointlyDV} respectively. 

\subsection{Instance-Level (Visual Only) Contrastive Learning}
\label{methods_visual_only_contrastive}
Instance-level contrastive learning methods are self-supervised methods that learn a representation of unlabeled data by discriminating between pairs of instances. They represent similar instances closer to each other in the embedding space, and dissimilar instances far apart. The quality of instance-level contrastive learning methods is influenced by the quality of data augmentations. This is the case because positive pairs for each instance are the augmented version of that instance, and the negative pairs are other instances in the training batch. To implement an instance-level contrastive learning method for video data, we follow the spatial augmentation process in the work of \citeauthor{CVRL} \cite{CVRL} that is consistent along the temporal dimension. Thus we will preserve the motion signal across frames, by generating random spatial augmentations across videos not across frames. We applied the following augmentations: random cropping with resizing, random horizontal flipping, color jitter, random grayscale, and Gaussian blur. Finally, we apply InfoNCE loss given as follows:
 \begin{equation} \label{simclr}
\ell_\text{NCE} = -\frac{1}{N}\sum_{i=1}^N \log \frac{\exp(sim(\mathbf{P}_{\textbf{v}}^i, \mathbf{P}_{\textbf{v}}^{'i}) / \tau)}{\sum^{2K}_{k=1} \mathbbold{1}_{[k\neq i]} \exp(sim(\mathbf{P}_\textbf{{v}}^{i}, \mathbf{P}_{\textbf{v}}^{k}) /\tau)} 
\end{equation}
\noindent
where $\tau$ is the positive temperature parameter, $\mathbf{P_v}$ is the representation of video frames, and $sim$ is a similarity function. By minimizing $\ell_\text{NCE}$ we are minimizing the distance between the instance $\mathbf{P_v}$, and its augmented versions $\mathbf{P_v}^{'}$, and maximize the distance between other instances. 

\subsection{Multi-Modal Contrastive Learning} 
\label{methods_multi_modal_contrastive}

To train a network that can project all modality inputs to the same embedding space, we learn a Masked Margin Softmax (MMS) loss \cite{MMS} between each input modality. Masked Margin Softmax (MMS) loss maximizes the similarity of correctly paired modalities and minimizes the similarity of incorrectly paired modalities. In the case of two modalities, text, and video frames. We calculate MMS loss as follows:

\begin{equation}
L_{MMS_{vw}} = L_{wv} + L_{vw} \label{loss_MMS_vw}
\end{equation} 

\begin{equation}
 L_{vw} = -\frac{1}{N}\sum_{i=1}^N \log \frac{{\rm e}^{\mathbf{P_{v_{i}}}\cdot \mathbf{P_{w_{i}}} - \delta } }{\rm e^{\mathbf{P_{v_{i}}}\cdot \mathbf{P_{w_{i}}} - \delta } + \sum^{N}_{j=1, j\neq i} \rm e^{\mathbf{P_{v_{j}}}\cdot \mathbf{P_{w_{i}}} }} \label{loss_vw} 
\end{equation}

\begin{equation}
 L_{wv} = -\frac{1}{N}\sum_{j=1}^N \log \frac{{\rm e}^{\mathbf{P_{v_{j}}}\cdot \mathbf{P_{w_{j}}} - \delta } }{\rm e^{\mathbf{P_{v_{j}}}\cdot \mathbf{P_{w_{j}}} - \delta} + \sum^{N}_{i=1, j\neq i} \rm e^{\mathbf{P_{v_{i}}}\cdot \mathbf{P_{w_{j}}} }} \label{loss_wv} 
\end{equation}

In the case of three modalities input, we calculate the MMS loss for each modality pair, and the final loss is the sum of all MMS losses between paired modalities.

\begin{equation}
L_{MMS} = L_{MMS_{vh}} + L_{MMS_{vw}} + L_{MMS_{hw}} \label{loss_MMS} 
\end{equation}

\subsection{Generative Self-Supervised Learning}
\label{generative_ssl}
In this work, we study the impact of the reconstruction objective function on the learned representation from unlabeled data. Either as the only objective function in the proxy task or as a part of multi-task self-supervised training. Since the training data is multi-modal, we follow a similar direction as \citeauthor{clusteringCont} \cite{clusteringCont} and perform the reconstruction for each modality individually. The final reconstruction loss is the sum of the reconstruction losses of all modalities. Thus we have 3 encoder-decoder models, $Q_f$, $Q_w$, and $Q_h$ for video, text, and audio respectively. Each decoder receives a feature representation $\mathbf{D}$ from the related reconstruction head. In the following you can see the reconstruction loss for the visual modality: 
\begin{equation}
L_{Rec_{\mathbf{v}}} = \text{MSE}(v, Q_v(\mathbf{D_v}))
\end{equation}\label{loss_rec_v}
The reconstruction loss is the mean-squared error (MSE) between the output features from the encoder-decoder model and the input features. Such loss penalizes the network for creating outputs (in this case features) different from the input features. The final reconstruction loss over the 3 modalities is given as: 
\begin{equation}
L_{Rec} = L_{Rec_{v}} + L_{Rec_{h}} + L_{Rec_{w}}
\end{equation}\label{loss_rec}

\subsection{Multi-Modal Contrastive Learning with Clustering}
\label{methods_multi_modal_contrastive_clustering}

This method learns a multi-modal representation utilizing the multi-modal contrastive objective function from \cref{methods_multi_modal_contrastive} that encourages modalities of the same instance represented closer to each other in the feature space. Moreover, it preserves the similarity between different instances by learning a multi-modal clustering objective function beside the multi-modal contrastive object function. The objective function that guides the multi-modal learning process consists of two losses: The first one is the multi-modal contrastive loss \cref{loss_MMS}, the second one is the multi-modal clustering loss that will be elaborated below.\\ 

\noindent
\textbf{Multi-modal clustering representation learning.} \\
To preserve the similarity between different modalities of similar instances, we formulate a clustering objective function that minimizes the distance between similar modalities. To create the clustering centroids we follow a similar direction of recent work \cite{clusteringCont} that creates a multi-modal centroid over joint multi-modal representations instead of creating a separate centroid for each modality. Thus, by using the clustering loss, we encourage the audio, visual, and text embeddings of similar videos (that share similar semantic information) to be represented closer to each other in the feature space. 

As shown in \cref{loss_main}, we compute the multi-modal representation $\mathbf{R_i}$ of each instance as the mean over the representations of each modality associated with that instance. As illustrated in \cref{fig:overview}, the clustering loss is performed over the output of the \textit{Clustering Head} which is $G$ (see \cref{methods_problem_formulation}). Assuming we have three modalities (video, text, and audio) as input then the multi-modal representation $\mathbf{R}$ for each instance $i$ is calculated as follows: \\

\begin{equation}
R_i (\mathbf{g}_{\mathbf{v}}^i,\mathbf{g}_{\mathbf{t}}^i,\mathbf{g}_{\mathbf{a}}^i)  =  (\mathbf{g}_{\mathbf{v}}^i+\mathbf{g}_{\mathbf{t}}^i+\mathbf{g}_{\mathbf{a}}^i) /3 \label{loss_main}
\end{equation}

After estimating the cluster centroids $(\mathbf{C}_1, \dotsc, \mathbf{C}_k)$, where $k$ is the number of clusters, we estimate the multi-modal centroids using K-means over the multi-modal representations $\mathbf{R}$ by minimizing the following equation:

\begin{equation}
\sum_{j=1}^{k}\sum_{i=1}^{n} \left | \mathbf{R}_{i}^{(j)} - \mathbf{C}_j\right |
\end{equation}
where $\mathbf{R}_{i}^{(j)}$ is the multimodal representation that belongs to cluster j with centroid $\mathbf{C}j$. Finally, we learn the Multi-modal clustering loss that minimizes the distance between the multi-modal representations $(\mathbf{R}_{1}^{(1)}, \dotsc, \mathbf{R}_{n}^{(k)})$ and the cluster centroids $(\mathbf{C}_1, \dotsc, \mathbf{C}_k)$. This loss updates all modality encoders to encourage modality embeddings to be presented closer to each other in the feature space. The multi-modal clustering loss encourages the multi-modal representations that belong to the same cluster to be closer to their centroid thus preserving the similarity between multimodal representations.
\begin{equation}
\ell_\text{Clu} = -\frac{1}{N}\sum_{i=1}^N \log \frac{{\rm e}^{\mathbf{R}_{i}^{j}\cdot \mathbf{C}_{j} - \delta } }{\sum^{K}_{k=1} \rm e^{\mathbf{R}_{i}^{j}\cdot \mathbf{C}_{k}}} 
\end{equation} \label{cluster_eq}

\subsection{Multi-Task Self-Supervised Learning}
\label{multi_task}

In this section, we analyze the ability to learn rich multi-modal representations in a multi-task self-supervised learning fashion. Here we will learn multiple self-supervised tasks simultaneously without supervision, only by leveraging the structure of the multi-modal data. The intuition behind multi-task self-supervised learning is to make the most of the structure of the large unlabeled datasets which leads to rich and better generalized representation learning. 
In this study, we investigate the following multi-modal task combination:
\begin{enumerate}
        \item \textbf{ConCluGen}: In this model, we jointly learn three self-supervised tasks. The first is multi-modal contrastive learning. The second is online multi-modal clustering. The third is multi-modal reconstruction. The objective function that guides the learning process in ConCluGen model is as follows:\\
        \begin{equation}
            L_{ConCluGen} = L_{MMS} + L_{Clu} + L_{Rec} \label{ConCluGen_eq}
        \end{equation} 
        \item \textbf{ConClu}: In this model, we jointly learn two self-supervised tasks. First is multi-modal contrastive learning. The second is online multi-modal clustering. The objective function that guides the learning process in ConClu model is as follows:\\
        \begin{equation}
            L_{ConClu} = L_{MMS} + L_{Clu} \label{ConClu_eq}
        \end{equation} 
        \item \textbf{ConGen}: In this model, we jointly learn two self-supervised tasks. First is multi-modal contrastive learning. The second is multi-modal reconstruction. The objective function that guides the learning process in ConGen model is as follows:\\
        \begin{equation}
            L_{ConGen} = L_{MMS} + L_{Rec} \label{ConGen_eq}
        \end{equation} 
\end{enumerate}

\section{Experiments and Analysis}

In this section, we evaluate the previously detailed self-supervised methods on three facial expression recognition datasets \cite{MOSEI,leeContextAwareEmotionRecognition2019,poriaMELDMultimodalMultiParty2019} . First, we present the
datasets used to evaluate our work. Then, we clarify the evaluation metrics (See the Appendix for implementation details). Finally, we analyze and discuss our results.

\subsection{Datasets}

\textbf{\gls{voxceleb2}.} We pretrain all methods on the large-scale face dataset \gls{voxceleb2} \cite{Chung18b}, which is \textit{not} annotated for \gls{fer}. \gls{voxceleb2} consists of 145,000 videos of celebrities and also contains audio and subtitles. Since the videos are already focused on the face, we did not perform any cropping.\\
For downstream task, we evaluate the \gls{caer} \cite{leeContextAwareEmotionRecognition2019}, \gls{meld} \cite{poriaMELDMultimodalMultiParty2019} and \gls{mosei} \cite{MOSEI} datasets. We employed a Pytorch MTCNN implementation\footnote{https://github.com/timesler/facenet-pytorch} to crop out the faces (see appendix for details).\\
\textbf{\gls{caer}.} \cite{leeContextAwareEmotionRecognition2019} consists of 13,000 videos, containing audio, of the TV series Friends. We automatically inferred subtitles from the audio data. Multiple speakers can be visible within one scene. The annotation labels are the 7 basic Ekman expressions \cite{ekmanUniversalsCulturalDifferences1971}.\\
\textbf{\gls{meld}.} \cite{poriaMELDMultimodalMultiParty2019} is a multi-modal dataset providing frames, audio, and subtitles and is also based on the TV series Friends, similarly to \gls{caer}. Its 13,000 videos feature single individuals only, cuts in a scene result in a different sample in this dataset. It provides 7 expression labels.\\
\textbf{\gls{mosei}.} \cite{MOSEI} contains 3,000 videos from YouTube in which people talk mostly directly into the camera. In contrast to \gls{caer} and \gls{meld}, \gls{mosei} offers smooth multi-label annotations for 6 emotions on a Likert scale from 0 to 3. We discretize the labels by setting every value greater than 0 to 1.\\
\textbf{Evaluation Metrics.} For the downstream task we measure \gls{wacc}, \gls{f1}, precision and recall, all weighted by the class occurences. \gls{caer}, \gls{meld} and \gls{mosei} are heavily imbalanced which makes weighting necessary.
\subsection{Evaluation Results}
In this section, we start with evaluating our model ConCluGen that we pretrained on \gls{voxceleb2} dataset against FER benchmarks and baselines. Then we conduct a detailed study to analyze the efficiency of the features obtained from the self-supervised methods mentioned above.
\subsubsection{Benchmarking Against SOTA}
In this section we are comparing the multi-task multi-modal self-supervised learning model \textbf{ConCluGen} that is pre-trained on VoxCeleb, to other multi-modal self-supervised benchmarks \cite{9206016,CAE-LR,Enrico}. Results in \cref{tab:MOSEI-SOTA} show that ConCluGen model pretrained on VoxCeleb and finetuned on \gls{mosei} dataset is outperforming CAE-LR \cite{CAE-LR}, which is a self-supervised model that is trained on \gls{mosei} dataset. It is also on par with \citeauthor{Enrico} \cite{Enrico} model performance, which uses more modalities (facial landmarks) than ConCluGen model for pretraining. It is worth mentioning that both CAE-LR \cite{CAE-LR}, and \citeauthor{Enrico} \cite{Enrico} are pretrained on \gls{mosei} and evaluate the downstream task on this dataset as well. However, ConCluGen model in \cref{tab:MOSEI-SOTA} is pretrained on \gls{voxceleb2}, and only finetuned on \gls{mosei}.

Moreover, our ConCluGen model, both with fintuning on the \gls{mosei} dataset on \cref{tab:MOSEI-SOTA} and without (\ie simple linear evaluation), are outperforming SSE-FT \cite{9206016}. SSE-FT is an independently pre-trained SSL model for vision, audio, and text modalities. They perform late fusion using an attention mechanism. Finally, they evaluate the model on \gls{mosei} dataset without finetuning.  

\noindent
Finally, results in \cref{tab:MOSEI-SOTA} show that ConCluGen model outperforms multiple fully supervised benchmarks \cite{CMIB,9431699,MOSEI,Chauhan2019ContextawareIA,DBLP:conf/naacl/DaiCLF21} and is on the par with \cite{DBLP:conf/naacl/DaiCLF21,Khare2020SelfSupervisedLW} on \gls{mosei}.

\begin{table}
  \centering
  \begin{tabular}{l cccc }
    \toprule
    Method & Year & Mod. & \gls{wacc} \\
    \midrule
    \multicolumn{4}{c}{Self-Supervised Baselines}  \\
    \midrule
    SSE-FT \cite{9206016} & 2020 & VAT & 55.7   \\
    CAE-LR \cite{CAE-LR} & 2021 & VAT & 61.03  \\
    \citeauthor{Enrico} \cite{Enrico} & 2022 & VATK & 66.70   \\
    ConCluGen (ours) & 2024 & VAT  & 66.48 \\
    \hline
    \multicolumn{4}{c}{Fully Supervised Baselines}  \\
    \hline
    CMIB \cite{CMIB} & 2022 & VAT & 48.2\\
    Huynh et al. \cite{9431699} & 2021 & VAT & 57.70 \\
    Graph-MFN \cite{MOSEI} & 2018 & VAT & 62.35 \\
    CIA \cite{Chauhan2019ContextawareIA} & 2019 & VAT & 62.88 \\
    MESM \cite{DBLP:conf/naacl/DaiCLF21} & 2021 & VAT & 66.80 \\
    \citeauthor{Khare2020SelfSupervisedLW} \cite{Khare2020SelfSupervisedLW}  & 2021 & VAT & 66.90 \\
    \citeauthor{WEN2021103178} \cite{WEN2021103178} & 2021 & VAT & 82.08 \\
    \citeauthor{shenoy-sardana-2020-multilogue} \cite{shenoy-sardana-2020-multilogue} & 2020 & VAT & 82.77  \\
    \bottomrule
  \end{tabular}
  \caption{Benchmark evaluations of our model against SOTA methods for CMU-MOSEI dataset. Modalities (\textit{Mod.}) are \textbf{V}ideo, \textbf{A}udio, \textbf{T}ext, (facial) \textbf{K}eypoints.}
  \label{tab:MOSEI-SOTA}
\end{table}

\begin{table}
  \centering
  \scalebox{0.85}{
  \begin{tabular}{l cc cc cc cc}
    \toprule
    Method & CMU-MOSEI  & CAER & MELD  \\
    \midrule
    Supervised baseline & 57.88 & 34.43 & 55.97  \\
    ConCluGen \small (linear eval.) & 60.0 & 37.5 & 56.6  \\
    ConCluGen \small (finetuned) & 66.48 & 50.75 & 60.03  \\  
    \bottomrule
  \end{tabular}}
  \caption{Comparison of the proposed method and baselines.}
  \label{tab:baselines}
\end{table}

\subsubsection{Supervised Baselines}
The supervised baseline shown in \cref{tab:baselines} consists of the same multi-model architecture as the ConCluGen model, but with a fully-supervised objective function on the target dataset.  Results in \cref{tab:baselines} show that both the linear evaluation and the finetuning approach outperform the fully supervised baseline for the three datasets. That highlights the efficiency of the multi-modal self-supervised learning method.

\begin{table}
  \centering
  \begin{tabular}{l cc cc cc cc}
    \toprule
    Method & \gls{wacc} & F1 & Prec. & Recall\\
    \midrule
    ConCluGen & 60.4 & 49.6 & 41.7 & 70.3 \\
    ConClu & 57.4 & 43.9 & 44.9 & 60.5 \\
    ConGen & 53.3 & 39.2 & 45.8 & 58.4 \\
    Multi-Cont & 56.4 & 44.4 & 43.9 & 63.1 \\
    Instance-Cont & 41.2 & 43.3 & 35.1 & 77.1 \\
    \hline
    Generative & 46.9 & 49.2 & 38 & 87.8 \\
    \hline
    ConCluGen/(vision) & 52.4 & 40.4 & 38.2 & 57.3 \\
    Multi-Cont \tiny Wout/text & 42.5 & 27.3 & 35.3 & 53.1 \\
    Multi-Cont \tiny Wout/audio  & 53.4 & 47.4 & 41.3 & 75.9 \\
    \bottomrule
  \end{tabular}
  \caption{Results of different combinations of self-supervised methods on CMU-MOSEI dataset. All metrics are weighted.}
  \label{tab:MOSEI}
\end{table}


\begin{table}
  \centering
  \begin{tabular}{l cc cc cc cc}
    \toprule
    Method & \gls{wacc} & F1 & Prec. & Recall\\
    \midrule
    ConCluGen & 56.6 & 53.1 & 58.5 & 56.6 \\
    ConClu & 58 & 54.7 & 57.6 & 58 \\
    ConGen & 57.4 & 52.7 & 57.1 & 57.4 \\
    Multi-Cont & 57.2 & 54.1 & 57.6 & 57.2 \\
    Instance-Cont & 46.3 & 32.1 & 25.3 & 46.3 \\
    \hline
    Generative & 52.1 & 40.9 & 37.5 & 52.1 \\
    \hline
put the RAVDESS stuff here292 292
Optimal Pretraining Hyperparameters Based on Downstream Accuracy
In the main section of this chapter, we followed a strict separation between pretraining and downstream293 293
training phase. This means that we optimized the hyperparameters for our pretraining solely on294 294
VoxCeleb2 and did not take the final classification accuracy on the downsream datasets into account.295 295
3.5 Discussion
• Add new part to discuss the difference between in-the-wild datasets and RAVDESS296 296
• Restate key findings297 297
    ConCluGen/(vision) & 31.0 & 26.3 & 28.7 & 31 \\
    Multi-Cont \tiny Wout/text &  48.1 & 31.4 & 24.3 & 48.1 \\
    Multi-Cont \tiny Wout/audio  & 54.3 & 46.9 & 58.2 & 54.3 \\
    \bottomrule
  \end{tabular}
  \caption{Results of different combinations of self-supervised methods on MELD dataset. All metrics are weighted.}
  \label{tab:MELD}
\end{table}

\begin{table}
  \centering
  \begin{tabular}{l cc cc cc cc}
    \toprule
    Method & \gls{wacc} & F1 & Prec. & Recall\\
    \midrule
    ConCluGen & 37.5 & 26.1 & 34.8 & 37.5 \\
    ConClu & 36.6 & 24 & 29.4 & 36.6 \\
    ConGen & 35.8 & 22.5 & 32.3 & 35.8 \\
    Multi-Cont & 34.6 & 22.3 & 30.7 & 34.6 \\
    Instance-Cont & 34.6 & 17.8 & 12 & 34.6 \\
    \hline
    Generative & 37.3 & 23.9 & 45.3 & 37.3 \\
    \hline
    ConCluGen/(vision) & 36.1 & 21.9 & 32.5 & 36.1 \\
    Multi-Cont \tiny Wout/text & 23.2 & 16.8 & 13.9 & 23.2 \\
    Multi-Cont \tiny Wout/audio  & 35.1 & 19.3 & 24 & 35.1 \\
    \bottomrule
  \end{tabular}
  \caption{Results of different combinations of self-supervised methods on CAER dataset. All metrics are weighted.}
  \label{tab:CARE}
\end{table}

\subsubsection{Evaluation of Self-Supervised Tasks}
We start with a comparison between individual self-supervised tasks, then we evaluate different permutations of combining different self-supervised tasks, all for facial expression recognition. \\

\begin{figure*}
  \centering
  \begin{subfigure}{0.24\linewidth}
       \includegraphics[width=\textwidth]{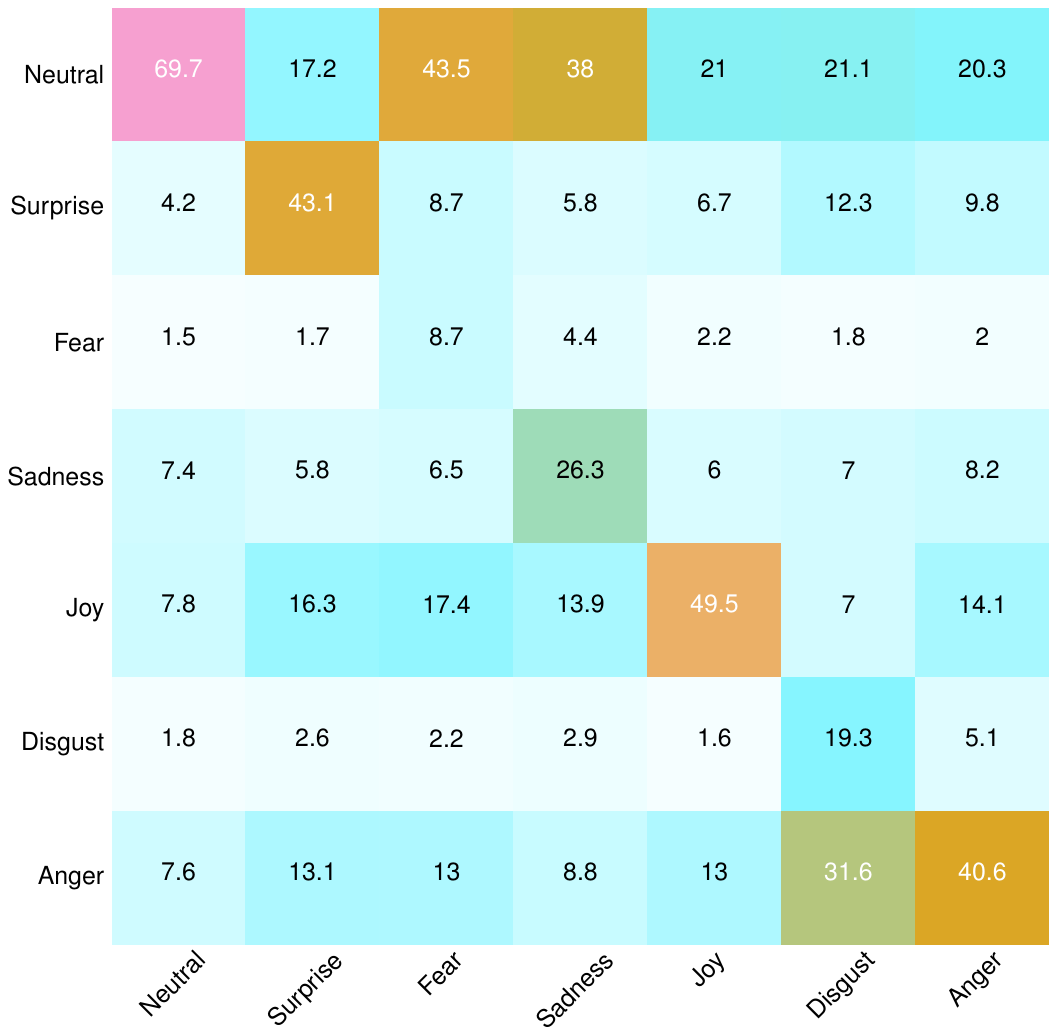}
    \caption{ConCluGen Model.}
    \label{fig:MELD-ConCluGen}
  \end{subfigure}
  \hfill
  \begin{subfigure}{0.24\linewidth}
         \includegraphics[width=\textwidth]{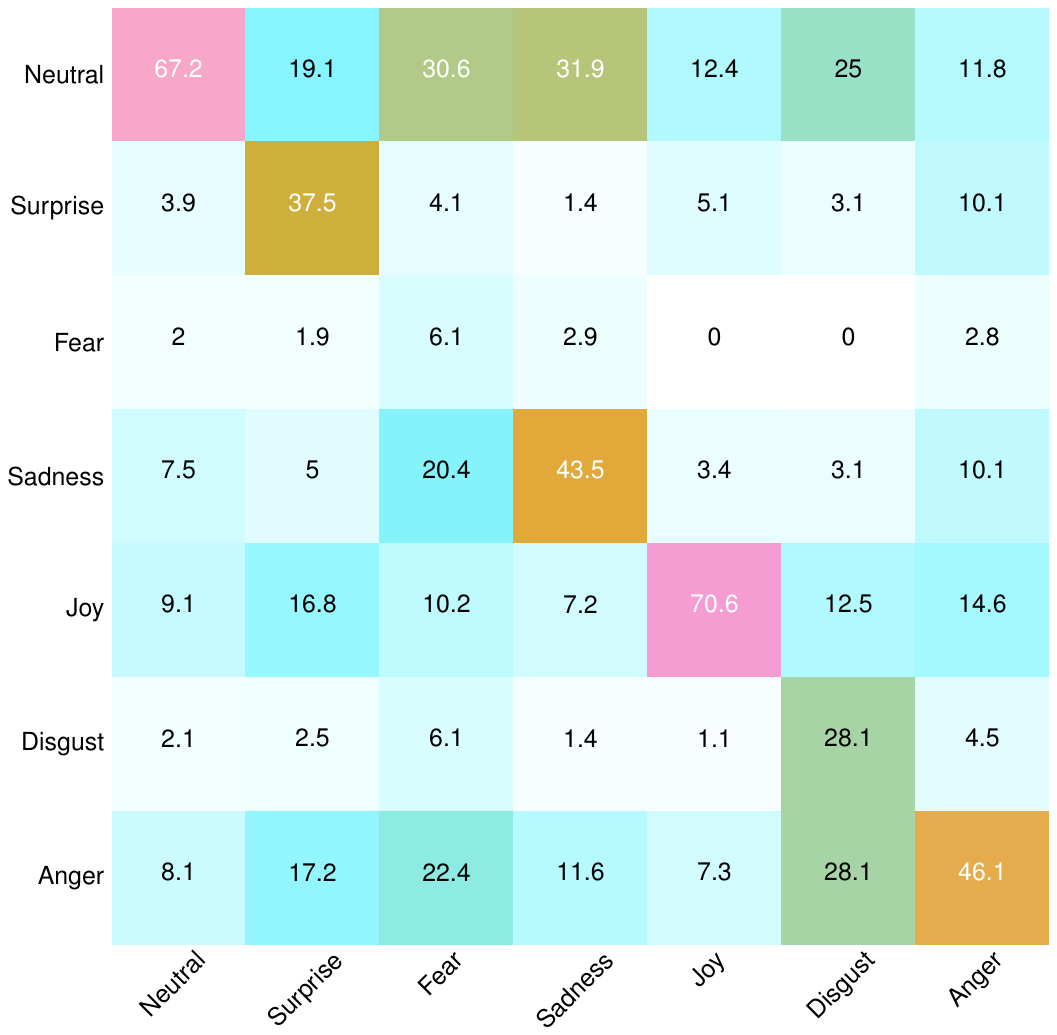}
        \caption{ConClu Model.}
        \label{fig:MELD-ConClu}
    \end{subfigure}
    \hfill
    \begin{subfigure}{0.24\linewidth}
    \includegraphics[width=\textwidth]{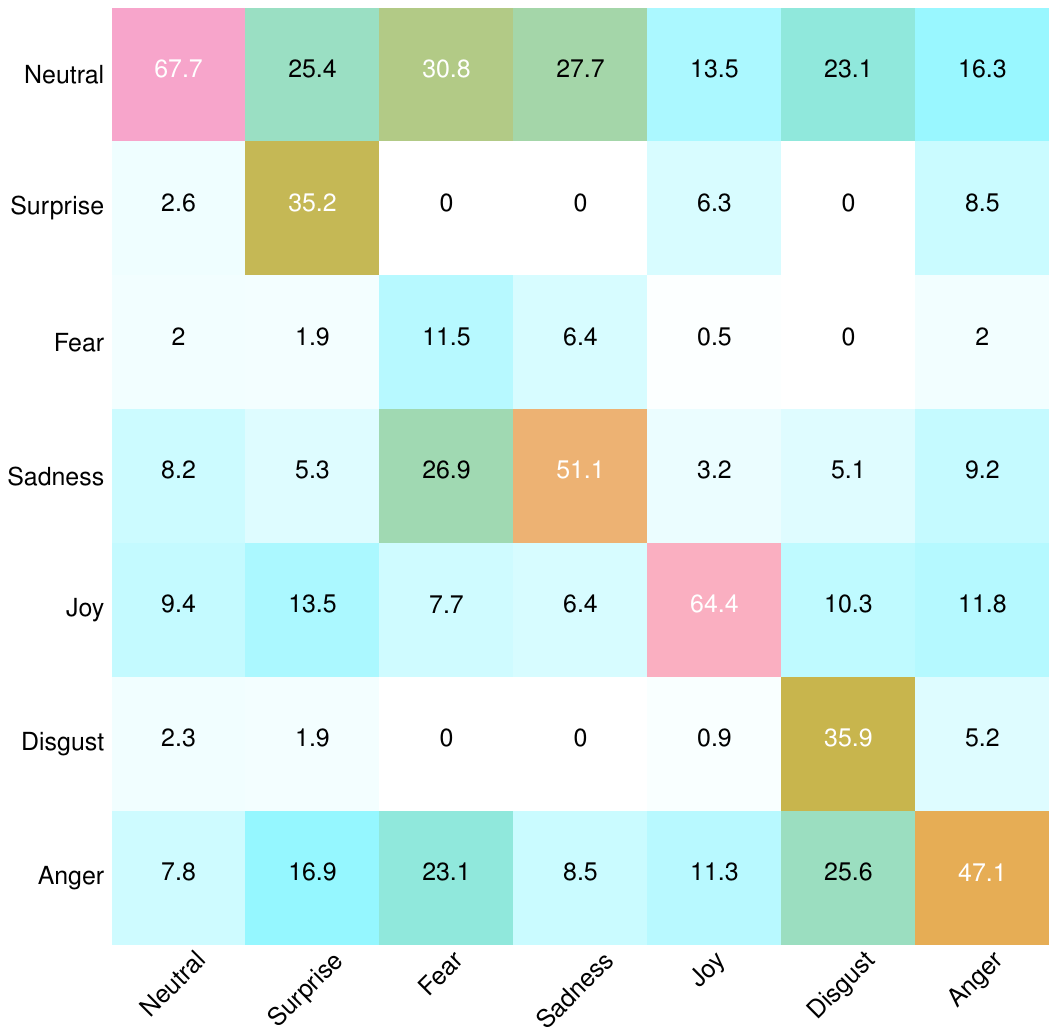}
    \caption{Multi-Cont Model.}
    \label{fig:MELD-Multi-Cont}
    \end{subfigure}
    \hfill
    \begin{subfigure}{0.24\linewidth}
       \includegraphics[width=\textwidth]{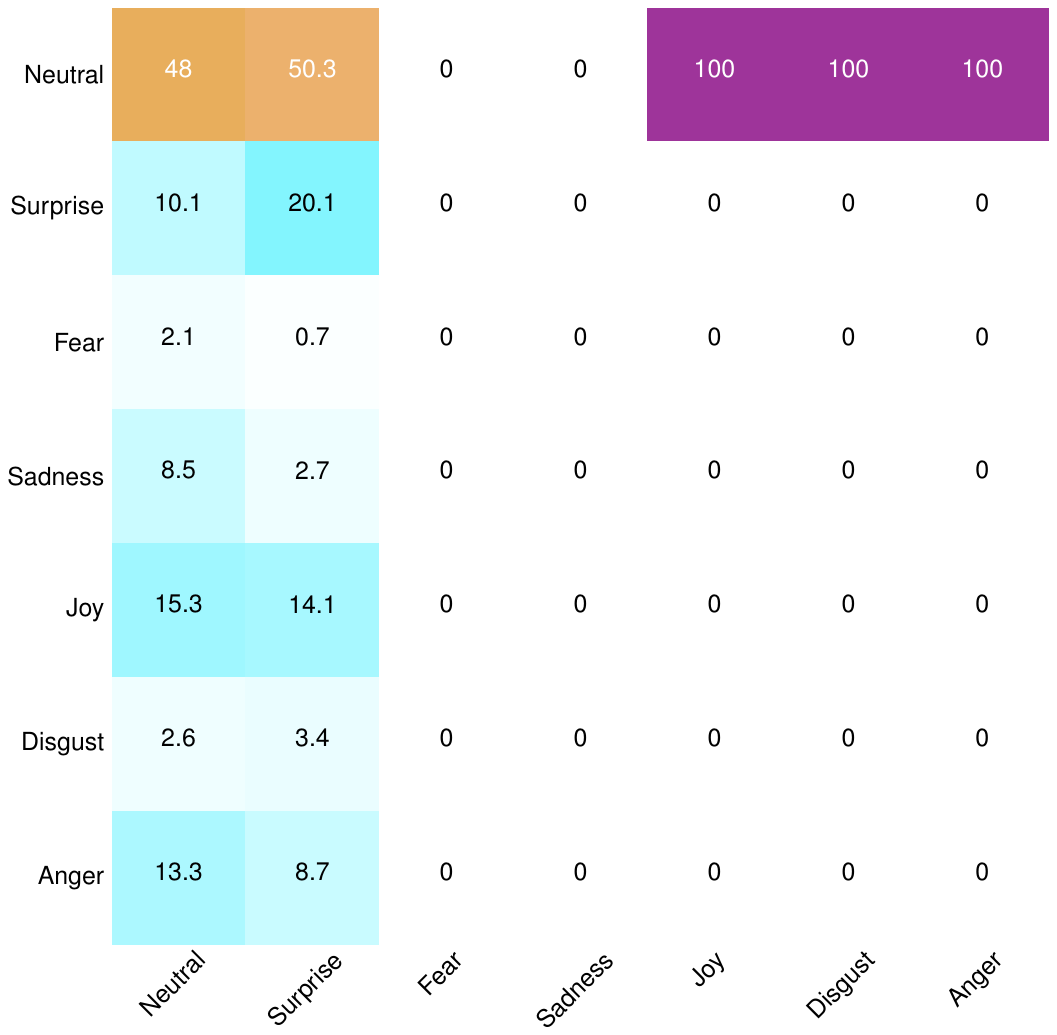}
    \caption{Instance-Cont Model.}
    \label{fig:MELD-Instance-Cont}
  \end{subfigure} 
  \caption{Confusion metrics for MELD dataset over different self-supervised models.}
  \label{fig:MELD_matrix}
\end{figure*}

\textbf{Instance-level contrastive learning vs.~multi-modal contrastive learning.}\\
Results in \cref{tab:MOSEI}, \cref{tab:MELD}, and \cref{tab:CARE} show that the multi-modal contrastive learning method \textbf{Multi-Cont} ( \cref{methods_multi_modal_contrastive}) outperforms the instance level contrastive learning \textbf{Instance-Cont} (\cref{methods_visual_only_contrastive}). 
Thus the self-supervised multi-modal contrastive method can learn better representations than the instance-level contrastive method (uni-modal). The confusion matrices for the \gls{meld} dataset in \cref{fig:MELD_matrix} indicate the quality of the classification performance of Multi-Cont model \cref{fig:MELD-Multi-Cont} across different emotion classes over Instance-Cont model \cref{fig:MELD-Instance-Cont}. See the Appendix for the confusion matrices of the \gls{caer} dataset.
Moreover, we conduct an experiment to evaluate which modality is more informative to be learned along with the visual modality. Results in  \cref{tab:MOSEI}, \cref{tab:MELD}, and \cref{tab:CARE} consistently show that the model can benefit more from text modality. 
\def\confmatsize{1.2}
\def\tsneconfmatsubfigsize{0.51}
\pgfplotsset{colormap/jet}

\textbf{Contrastive learning with clustering.}\\
Results on \cref{tab:MELD}, and \cref{tab:CARE} show that learning online multi-modal clustering task along with multi-modal contrastive task during the pre-training phase, such as in \textbf{ConClu} model (\cref{methods_multi_modal_contrastive_clustering}), improves the results on the downstream task of \gls{fer} for both \gls{meld} and \gls{caer}. \cref{tab:MOSEI} shows that the results are comparable between both models on \gls{mosei} dataset. The hypothesis behind such a model is that by supporting the contrastive method with distance-based clustering, we can better capture semantic structures from the data. \cref{fig:MELD-ConClu} shows how the quality of the classification in ConClu model enhanced over emotion classes on \gls{meld} dataset over Multi-Cont model (\cref{fig:MELD-Multi-Cont}).\\

\textbf{Generative self-supervised learning vs.~contrastive self-supervised learning.}\\
To answer this question we compare the generative multi-modal model \textbf{Generative} (\cref{generative_ssl}) that reconstructs each modality representation separately, to both multi-modal contrastive learning method \textbf{Multi-Cont} and instance-level contrastive \textbf{Instance-Cont}. Results on \gls{mosei} and \gls{caer} datasets in \cref{tab:MOSEI} and \cref{tab:CARE} show that the generative loss is a robust objective function as a proxy task for \gls{fer}. On the other hand, results for \gls{meld} dataset in \cref{tab:MELD} show that the  \textbf{Generative} model is outperforming \textbf{Instance-Cont} but not the \textbf{Multi-Cont} model.

\textbf{Multi-task multi-modal self-supervised learning.}\\
In the previous section, results showed how powerful the generative objective function is as a proxy task compared to contrastive objective functions on some datasets. On the other hand, results in \cref{tab:MOSEI}, \cref{tab:MELD}, \cref{tab:CARE}, show that for all datasets, minimizing a reconstruction loss along with a multi-modal contrastive loss \textbf{ConGen} leads to models that underperform the model that we pre-trained using only the multi-modal contrastive objective function. \\
Finally, we evaluate the model that learns using the multi-task multi-modal self-supervised objective function \textbf{ConCluGen} \cref{multi_task}. Results in  \cref{tab:MOSEI}, and \cref{tab:CARE}, show that for all datasets except for \gls{meld} \cref{tab:MELD}, ConCluGen model is outperforming all other models. \\

\textbf{How informative is the individual modality representation when learning the multi-modal multi-task self-supervised model \textbf{ConCluGen}?}\\
To answer this question we evaluate the results using only the visual representation from the multi-modal multi-task model \textbf{ConCluGen} to the results of \textbf{Instance-Cont} model that only utilizes the vision modality during pretraining. \cref{tab:CARE}, and \cref{tab:MOSEI}, show that \textbf{ConCluGen/(vision)} is outperforming the model that we pre-trained using the instance level contrastive method. On the other hand, results in \cref{tab:MELD} show that we need to utilize the three modalities that are learned by \textbf{ConCluGen} model, to outperform the \textbf{Instance-Cont} model.
\section{Conclusion and Future Work}
In this work, we employed a multi-task multi-modal self-supervised method for facial expression recognition on three different in-the-wild datasets. Our \textbf{ConCluGen} model outperforms several multi-modal self-supervised as well as supervised baseline models. We conduct an extensive experimental study to evaluate the performance of pre-trained models with multiple self-supervised tasks. 

First, we assess the gains obtained by including multiple data modalities in the self-supervised algorithm, by evaluating the performance of the model pre-trained with a multi-modal contrastive method against a model pre-trained using a contrastive method with the visual modality only, following instance-based state-of-the-art self-supervised algorithms. We found that the multi-modal contrastive method learns more informative representations than the instance-based contrastive (uni-modal method), \ie the resulting model from the multi-modal method outperforms the uni-modal model on the facial expression recognition task. In another related experiment, we discarded the other data modalities from the multi-modal model, and only kept the visual modality part of the model. In this experiment, we found that even only the visual part of the multi-modal still outperforms the instance-based uni-modal model which was trained on the visual modality only. These experiments illustrate the gains obtained when including other data modalities in the self-supervised pre-training stage; the resulting features contain more information about the data - facial expressions in our case. In addition, another noteworthy observation here is that the text modality seems to enrich the resulting representations with more information about facial expressions than the audio modality.

Moreover, in another set of experiments, we evaluate the performance gains when pretraining a model with a self-supervised task that combines both a multi-modal contrastive loss and a multi-modal clustering loss simultaneously. Here, we found that combining a contrastive loss with a distance-based clustering loss encourages the model to learn more semantic structure from the data. This effect is demonstrated clearly in the improved downstream \gls{fer} performance. We believe that employing a clustering loss in combination with a contrastive loss mitigates the issue of class collision that contrastive learning methods alone could encounter \cite{classcollision}, as explained earlier.

Finally, it is noteworthy that performing the self-supervised pre-training phase in a multi-task fashion captures more semantically meaningful representations than performing each task alone. In other words, for a challenging downstream task, such as facial expression recognition in the wild, the representations learned by each individual self-supervised task alone do not seem sufficient, as demonstrated by the superior performance of \textbf{ConCluGen} against all other self-supervised models.

In future work, we would like to expand the multi-task multi-modal method for more modalities such as facial landmarks. Additionally, we aim to evaluate our multi-task multi-modal \textbf{ConCluGen} on different downstream tasks, such as facial action unit detection, face detection, and sentiment analysis. The code and the pre-trained models implemented in this work are publicly available for the research community as baselines for future studies \footnote{https://github.com/tub-cv-group/conclugen}.

\section*{Acknowledgements}

Funded by the Deutsche Forschungsgemeinschaft (DFG, German Research Foundation) under Germany’s Excellence Strategy – EXC 2002/1 “Science of Intelligence” – project number 390523135. In addition, this work has been partially supported by MIAI@Grenoble Alpes, (ANR-19-P3IA-0003)

{
    \small
    \bibliographystyle{ieeenat_fullname}
    \bibliography{main}

\begin{thebibliography}{60}
\providecommand{\natexlab}[1]{#1}
\providecommand{\url}[1]{\texttt{#1}}
\expandafter\ifx\csname urlstyle\endcsname\relax
  \providecommand{\doi}[1]{doi: #1}\else
  \providecommand{\doi}{doi: \begingroup \urlstyle{rm}\Url}\fi

\bibitem[Abdat et~al.(2011)Abdat, Maaoui, and Pruski]{6131215}
F. Abdat, C. Maaoui, and A. Pruski.
\newblock Human-computer interaction using emotion recognition from facial expression.
\newblock In \emph{2011 UKSim 5th European Symposium on Computer Modeling and Simulation}, pages 196--201, 2011.

\bibitem[Alwassel et~al.(2020)Alwassel, Mahajan, Korbar, Torresani, Ghanem, and Tran]{10.5555/3495724.3496542}
Humam Alwassel, Dhruv Mahajan, Bruno Korbar, Lorenzo Torresani, Bernard Ghanem, and Du Tran.
\newblock Self-supervised learning by cross-modal audio-video clustering.
\newblock In \emph{Proceedings of the 34th International Conference on Neural Information Processing Systems}, Red Hook, NY, USA, 2020. Curran Associates Inc.

\bibitem[Andonian(2024)]{pretorchedx}
Alex Andonian.
\newblock Pretorched-x.
\newblock \url{https://github.com/alexandonian/pretorched-x}, 2024.

\bibitem[Asano et~al.(2020)Asano, Patrick, 0001, and Vedaldi]{AsanoP0V20}
Yuki~Markus Asano, Mandela Patrick, Christian~Rupprecht 0001, and Andrea Vedaldi.
\newblock Labelling unlabelled videos from scratch with multi-modal self-supervision.
\newblock In \emph{Advances in Neural Information Processing Systems 33: Annual Conference on Neural Information Processing Systems 2020, NeurIPS 2020, December 6-12, 2020, virtual}, 2020.

\bibitem[Aviezer et~al.(2017)Aviezer, Ensenberg, and Hassin]{aviezerInherentlyContextualizedNature2017}
Hillel Aviezer, Noga Ensenberg, and Ran~R Hassin.
\newblock The inherently contextualized nature of facial emotion perception.
\newblock \emph{Current Opinion in Psychology}, 17:\penalty0 47--54, 2017.

\bibitem[Bagher~Zadeh et~al.(2018)Bagher~Zadeh, Liang, Poria, Cambria, and Morency]{MOSEI}
AmirAli Bagher~Zadeh, Paul~Pu Liang, Soujanya Poria, Erik Cambria, and Louis-Philippe Morency.
\newblock Multimodal language analysis in the wild: {CMU}-{MOSEI} dataset and interpretable dynamic fusion graph.
\newblock In \emph{Proceedings of the 56th Annual Meeting of the Association for Computational Linguistics (Volume 1: Long Papers)}, pages 2236--2246, Melbourne, Australia, 2018. Association for Computational Linguistics.

\bibitem[Baum et~al.(2020)Baum, Rabovsky, Rose, and Abdel~Rahman]{baumClearJudgmentsBased2020}
Julia Baum, Milena Rabovsky, Sebastian~Benjamin Rose, and Rasha Abdel~Rahman.
\newblock Clear judgments based on unclear evidence: {{Person}} evaluation is strongly influenced by untrustworthy gossip.
\newblock \emph{Emotion}, 20\penalty0 (2):\penalty0 248--260, 2020.

\bibitem[Chauhan et~al.(2019)Chauhan, Akhtar, Ekbal, and Bhattacharyya]{Chauhan2019ContextawareIA}
Dushyant~Singh Chauhan, Md.~Shad Akhtar, Asif Ekbal, and Pushpak Bhattacharyya.
\newblock Context-aware interactive attention for multi-modal sentiment and emotion analysis.
\newblock In \emph{Conference on Empirical Methods in Natural Language Processing}, 2019.

\bibitem[Chen et~al.(2021)Chen, Rouditchenko, Duarte, Kuehne, Thomas, Boggust, Panda, Kingsbury, Feris, Harwath, Glass, Picheny, and Chang]{clusteringCont}
Brian Chen, Andrew Rouditchenko, Kevin Duarte, Hilde Kuehne, Samuel Thomas, Angie Boggust, Rameswar Panda, Brian Kingsbury, Rog{\'e}rio~Schmidt Feris, David~F. Harwath, James~R. Glass, Michael Picheny, and Shih-Fu Chang.
\newblock Multimodal clustering networks for self-supervised learning from unlabeled videos.
\newblock \emph{2021 IEEE/CVF International Conference on Computer Vision (ICCV)}, pages 7992--8001, 2021.

\bibitem[Chen et~al.(2020)Chen, Kornblith, Norouzi, and Hinton]{simclr}
Ting Chen, Simon Kornblith, Mohammad Norouzi, and Geoffrey Hinton.
\newblock A simple framework for contrastive learning of visual representations.
\newblock \emph{arXiv preprint arXiv:2002.05709}, 2020.

\bibitem[Chung et~al.(2018)Chung, Nagrani, and Zisserman]{Chung18b}
J.~S. Chung, A. Nagrani, and A. Zisserman.
\newblock {{VoxCeleb2}}: {{Deep}} speaker recognition.
\newblock In \emph{{{INTERSPEECH}}}, 2018.

\bibitem[Dai et~al.(2021)Dai, Cahyawijaya, Liu, and Fung]{DBLP:conf/naacl/DaiCLF21}
Wenliang Dai, Samuel Cahyawijaya, Zihan Liu, and Pascale Fung.
\newblock Multimodal end-to-end sparse model for emotion recognition.
\newblock In \emph{Proceedings of the 2021 Conference of the North American Chapter of the Association for Computational Linguistics: Human Language Technologies, {NAACL-HLT} 2021, Online, June 6-11, 2021}, pages 5305--5316. Association for Computational Linguistics, 2021.

\bibitem[Deng et~al.(2009)Deng, Dong, Socher, Li, Li, and {Fei-Fei}]{dengImageNetLargescaleHierarchical2009}
Jia Deng, Wei Dong, Richard Socher, Li-Jia Li, Kai Li, and Li {Fei-Fei}.
\newblock {{ImageNet}}: {{A}} large-scale hierarchical image database.
\newblock In \emph{2009 {{IEEE Conference}} on {{Computer Vision}} and {{Pattern Recognition}}}, pages 248--255, 2009.

\bibitem[Devlin et~al.(2019)Devlin, Chang, Lee, and Toutanova]{bert}
Jacob Devlin, Ming-Wei Chang, Kenton Lee, and Kristina Toutanova.
\newblock {BERT}: Pre-training of deep bidirectional transformers for language understanding.
\newblock In \emph{Proceedings of the 2019 Conference of the North {A}merican Chapter of the Association for Computational Linguistics: Human Language Technologies, Volume 1 (Long and Short Papers)}, pages 4171--4186, Minneapolis, Minnesota, 2019. Association for Computational Linguistics.

\bibitem[Doersch and Zisserman(2017)]{Doersch2017MultitaskSV}
Carl Doersch and Andrew Zisserman.
\newblock Multi-task self-supervised visual learning.
\newblock \emph{2017 IEEE International Conference on Computer Vision (ICCV)}, pages 2070--2079, 2017.

\bibitem[Eiserbeck et~al.(2023)Eiserbeck, Maier, Baum, and Abdel~Rahman]{eiserbeckDeepfakeSmilesMatter2023}
Anna Eiserbeck, Martin Maier, Julia Baum, and Rasha Abdel~Rahman.
\newblock Deepfake smiles matter less---the psychological and neural impact of presumed {{AI-generated}} faces.
\newblock \emph{Scientific Reports}, 13\penalty0 (1):\penalty0 16111, 2023.

\bibitem[Ekman(1971)]{ekmanUniversalsCulturalDifferences1971}
Paul Ekman.
\newblock Universals and cultural differences in facial expressions of emotion.
\newblock \emph{Nebraska Symposium on Motivation}, 19:\penalty0 207--283, 1971.

\bibitem[Feldman~Barrett and Kensinger(2010)]{feldmanbarrettContextRoutinelyEncoded2010}
Lisa Feldman~Barrett and E. Kensinger.
\newblock Context {{Is Routinely Encoded During Emotion Perception}}.
\newblock \emph{Psychological science}, 2010.

\bibitem[Franceschini et~al.(2022)Franceschini, Fini, Beyan, Conti, Arrigoni, and Ricci]{Enrico}
R. Franceschini, E. Fini, C. Beyan, A. Conti, F. Arrigoni, and E. Ricci.
\newblock Multimodal emotion recognition with modality-pairwise unsupervised contrastive loss.
\newblock In \emph{2022 26th International Conference on Pattern Recognition (ICPR)}, pages 2589--2596, Los Alamitos, CA, USA, 2022. IEEE Computer Society.

\bibitem[Gidaris et~al.(2018)Gidaris, Singh, and Komodakis]{conf/iclr/GidarisSK18}
Spyros Gidaris, Praveer Singh, and Nikos Komodakis.
\newblock Unsupervised representation learning by predicting image rotations.
\newblock In \emph{6th International Conference on Learning Representations, {ICLR} 2018, Vancouver, BC, Canada, April 30 - May 3, 2018, Conference Track Proceedings}. OpenReview.net, 2018.

\bibitem[Halawa et~al.(2020)Halawa, W{\"o}llhaf, Vellasques, Sanz, and Hellwich]{halawa2020learning}
Marah Halawa, Manuel W{\"o}llhaf, Eduardo Vellasques, Urko~S{\'a}nchez Sanz, and Olaf Hellwich.
\newblock Learning disentangled expression representations from facial images.
\newblock \emph{arXiv preprint arXiv:2008.07001}, 2020.

\bibitem[Halawa et~al.(2022)Halawa, Hellwich, and Bideau]{ABC}
Marah Halawa, Olaf Hellwich, and Pia Bideau.
\newblock Action-based contrastive learning for trajectory prediction.
\newblock In \emph{Computer Vision -- ECCV 2022}, pages 143--159, Cham, 2022. Springer Nature Switzerland.

\bibitem[Harwath et~al.(2018)Harwath, Recasens, Sur{\'i}s, Chuang, Torralba, and Glass]{Harwath2018JointlyDV}
David~F. Harwath, Adri{\`a} Recasens, D{\'i}dac Sur{\'i}s, Galen Chuang, Antonio Torralba, and James~R. Glass.
\newblock Jointly discovering visual objects and spoken words from raw sensory input.
\newblock \emph{International Journal of Computer Vision}, 128:\penalty0 620 -- 641, 2018.

\bibitem[He et~al.(2015)He, Zhang, Ren, and Sun]{He2015DeepRL}
Kaiming He, X. Zhang, Shaoqing Ren, and Jian Sun.
\newblock Deep residual learning for image recognition.
\newblock \emph{2016 IEEE Conference on Computer Vision and Pattern Recognition (CVPR)}, pages 770--778, 2015.

\bibitem[Huynh et~al.(2021)Huynh, Yang, Lee, and Kim]{9431699}
Van~Thong Huynh, Hyung-Jeong Yang, Guee-Sang Lee, and Soo-Hyung Kim.
\newblock End-to-end learning for multimodal emotion recognition in video with adaptive loss.
\newblock \emph{IEEE MultiMedia}, 28\penalty0 (2):\penalty0 59--66, 2021.

\bibitem[Ilharco et~al.(2019)Ilharco, Zhang, and Baldridge]{MMS}
Gabriel Ilharco, Yuan Zhang, and Jason Baldridge.
\newblock Large-scale representation learning from visually grounded untranscribed speech.
\newblock In \emph{Proceedings of the 23rd Conference on Computational Natural Language Learning (CoNLL)}, pages 55--65, Hong Kong, China, 2019. Association for Computational Linguistics.

\bibitem[Kay et~al.(2017)Kay, Carreira, Simonyan, Zhang, Hillier, Vijayanarasimhan, Viola, Green, Back, Natsev, Suleyman, and Zisserman]{kinetics}
Will Kay, Jo{\~{a}}o Carreira, Karen Simonyan, Brian Zhang, Chloe Hillier, Sudheendra Vijayanarasimhan, Fabio Viola, Tim Green, Trevor Back, Paul Natsev, Mustafa Suleyman, and Andrew Zisserman.
\newblock The kinetics human action video dataset.
\newblock \emph{CoRR}, abs/1705.06950, 2017.

\bibitem[Khare et~al.(2020)Khare, Parthasarathy, and Sundaram]{Khare2020SelfSupervisedLW}
Aparna Khare, Srinivas Parthasarathy, and Shiva Sundaram.
\newblock Self-supervised learning with cross-modal transformers for emotion recognition.
\newblock \emph{2021 IEEE Spoken Language Technology Workshop (SLT)}, pages 381--388, 2020.

\bibitem[Kim et~al.(2024)Kim, Adaloglou, Osterland, Morelli, Halawa, K{\"o}nig, Gnutt, and Zapata]{Kim2023.04.28.538691}
Vladislav Kim, Nikolaos Adaloglou, Marc Osterland, Flavio~M. Morelli, Marah Halawa, Tim K{\"o}nig, David Gnutt, and Paula A.~Marin Zapata.
\newblock Self-supervision advances morphological profiling by unlocking powerful image representations.
\newblock \emph{bioRxiv}, 2024.

\bibitem[Koromilas and Giannakopoulos(2021)]{CAE-LR}
Panagiotis Koromilas and Theodoros Giannakopoulos.
\newblock Unsupervised multimodal language representations using convolutional autoencoders.
\newblock \emph{CoRR}, abs/2110.03007, 2021.

\bibitem[Kosti et~al.(2019)Kosti, Alvarez, Recasens, and Lapedriza]{kostiContextBasedEmotion2019a}
Ronak Kosti, Jose Alvarez, Adria Recasens, and Agata Lapedriza.
\newblock Context {{Based Emotion Recognition}} using {{EMOTIC Dataset}}.
\newblock \emph{IEEE Transactions on Pattern Analysis and Machine Intelligence}, pages 1--1, 2019.

\bibitem[Larsson et~al.(2017)Larsson, Maire, and Shakhnarovich]{Larsson_2017}
Gustav Larsson, Michael Maire, and Gregory Shakhnarovich.
\newblock Colorization as a proxy task for visual understanding.
\newblock In \emph{2017 IEEE Conference on Computer Vision and Pattern Recognition (CVPR)}. IEEE, 2017.

\bibitem[Lee et~al.(2019)Lee, Kim, Kim, Park, and Sohn]{leeContextAwareEmotionRecognition2019}
Jiyoung Lee, Seungryong Kim, Sunok Kim, Jungin Park, and Kwanghoon Sohn.
\newblock Context-{{Aware Emotion Recognition Networks}}.
\newblock In \emph{2019 {{IEEE}}/{{CVF International Conference}} on {{Computer Vision}} ({{ICCV}})}, pages 10142--10151, 2019.

\bibitem[Li and Deng(2022)]{survayFER}
Shan Li and Weihong Deng.
\newblock Deep facial expression recognition: A survey.
\newblock \emph{IEEE Transactions on Affective Computing}, 13\penalty0 (3):\penalty0 1195--1215, 2022.

\bibitem[Liu et~al.(2018)Liu, Wei, Shao, Sheng, Yan, and Wang]{Liu2018ExploringDF}
Yu Liu, Fangyin Wei, Jing Shao, Lu Sheng, Junjie Yan, and Xiaogang Wang.
\newblock Exploring disentangled feature representation beyond face identification.
\newblock \emph{2018 IEEE/CVF Conference on Computer Vision and Pattern Recognition}, pages 2080--2089, 2018.

\bibitem[Loshchilov and Hutter(2017{\natexlab{a}})]{loshchilovDecoupledWeightDecay2017}
I. Loshchilov and F. Hutter.
\newblock Decoupled {{Weight Decay Regularization}}.
\newblock In \emph{International {{Conference}} on {{Learning Representations}}}, 2017{\natexlab{a}}.

\bibitem[Loshchilov and Hutter(2017{\natexlab{b}})]{loshchilovSGDRSTOCHASTICGRADIENT2017}
Ilya Loshchilov and Frank Hutter.
\newblock {SGDR:} stochastic gradient descent with warm restarts.
\newblock In \emph{5th International Conference on Learning Representations, {ICLR} 2017, Toulon, France, April 24-26, 2017, Conference Track Proceedings}. OpenReview.net, 2017{\natexlab{b}}.

\bibitem[Mai et~al.(2022)Mai, Zeng, and Hu]{CMIB}
Sijie Mai, Ying Zeng, and Haifeng Hu.
\newblock Multimodal information bottleneck: Learning minimal sufficient unimodal and multimodal representations.
\newblock \emph{IEEE Transactions on Multimedia}, pages 1--1, 2022.

\bibitem[Maier et~al.(2022)Maier, Blume, Bideau, Hellwich, and Abdel~Rahman]{maierKnowledgeaugmentedFacePerception2022}
Martin Maier, Florian Blume, Pia Bideau, Olaf Hellwich, and Rasha Abdel~Rahman.
\newblock Knowledge-augmented face perception: {{Prospects}} for the {{Bayesian}} brain-framework to align {{AI}} and human vision.
\newblock \emph{Consciousness and Cognition}, 101:\penalty0 103301, 2022.

\bibitem[Mikolov et~al.(2013)Mikolov, Chen, Corrado, and Dean]{word2vec}
Tomas Mikolov, Kai Chen, Gregory~S. Corrado, and Jeffrey Dean.
\newblock Efficient estimation of word representations in vector space.
\newblock In \emph{International Conference on Learning Representations}, 2013.

\bibitem[Noroozi and Favaro(2016)]{10.1007/978-3-319-46466-4_5}
Mehdi Noroozi and Paolo Favaro.
\newblock Unsupervised learning of visual representations by solving jigsaw puzzles.
\newblock In \emph{Computer Vision -- ECCV 2016}, pages 69--84, Cham, 2016. Springer International Publishing.

\bibitem[Owens and Efros(2018)]{Owens2018AudioVisualSA}
Andrew Owens and Alexei~A. Efros.
\newblock Audio-visual scene analysis with self-supervised multisensory features.
\newblock In \emph{European Conference on Computer Vision}, 2018.

\bibitem[Pathak et~al.(2019)Pathak, Gandhi, and Gupta]{pathak19disagreement}
Deepak Pathak, Dhiraj Gandhi, and Abhinav Gupta.
\newblock Self-supervised exploration via disagreement.
\newblock In \emph{ICML}, 2019.

\bibitem[Poria et~al.(2019)Poria, Hazarika, Majumder, Naik, Cambria, and Mihalcea]{poriaMELDMultimodalMultiParty2019}
Soujanya Poria, Devamanyu Hazarika, Navonil Majumder, Gautam Naik, Erik Cambria, and Rada Mihalcea.
\newblock {{MELD}}: {{A Multimodal Multi-Party Dataset}} for {{Emotion Recognition}} in {{Conversations}}.
\newblock \emph{Proceedings of the 57th Annual Meeting of the Association for Computational Linguistics}, pages 527--536, 2019.

\bibitem[Qian et~al.(2021)Qian, Meng, Gong, Yang, Wang, Belongie, and Cui]{CVRL}
R. Qian, T. Meng, B. Gong, M. Yang, H. Wang, S. Belongie, and Y. Cui.
\newblock Spatiotemporal contrastive video representation learning.
\newblock In \emph{2021 IEEE/CVF Conference on Computer Vision and Pattern Recognition (CVPR)}, pages 6960--6970, Los Alamitos, CA, USA, 2021. IEEE Computer Society.

\bibitem[Radford et~al.(2021)Radford, Kim, Hallacy, Ramesh, Goh, Agarwal, Sastry, Askell, Mishkin, Clark, Krueger, and Sutskever]{CLIP}
Alec Radford, Jong~Wook Kim, Chris Hallacy, Aditya Ramesh, Gabriel Goh, Sandhini Agarwal, Girish Sastry, Amanda Askell, Pamela Mishkin, Jack Clark, Gretchen Krueger, and Ilya Sutskever.
\newblock Learning transferable visual models from natural language supervision.
\newblock In \emph{Proceedings of the 38th International Conference on Machine Learning, {ICML} 2021, 18-24 July 2021, Virtual Event}, pages 8748--8763. {PMLR}, 2021.

\bibitem[Russell(1997)]{russellReadingEmotionsFaces1997}
James~A. Russell.
\newblock Reading emotions from and into faces: {{Resurrecting}} a dimensional-contextual perspective.
\newblock In \emph{The {{Psychology}} of {{Facial Expression}}}, pages 295--320. Cambridge University Press, Cambridge, 1997.

\bibitem[Sanh et~al.(2019)Sanh, Debut, Chaumond, and Wolf]{Sanh2019DistilBERTAD}
Victor Sanh, Lysandre Debut, Julien Chaumond, and Thomas Wolf.
\newblock Distilbert, a distilled version of bert: smaller, faster, cheaper and lighter.
\newblock \emph{ArXiv}, abs/1910.01108, 2019.

\bibitem[Schiappa et~al.(2022)Schiappa, Rawat, and Shah]{ChantrySchiappa2022SelfSupervisedLF}
Madeline~Chantry Schiappa, Yogesh~Singh Rawat, and Mubarak Shah.
\newblock Self-supervised learning for videos: A survey.
\newblock \emph{ACM Computing Surveys}, 55:\penalty0 1 -- 37, 2022.

\bibitem[Shenoy and Sardana(2020)]{shenoy-sardana-2020-multilogue}
Aman Shenoy and Ashish Sardana.
\newblock Multilogue-net: A context-aware {RNN} for multi-modal emotion detection and sentiment analysis in conversation.
\newblock In \emph{Second Grand-Challenge and Workshop on Multimodal Language (Challenge-HML)}, pages 19--28, Seattle, USA, 2020. Association for Computational Linguistics.

\bibitem[Shu et~al.(2022)Shu, Gu, Yang, and Lo]{Shu2022RevisitingSC}
Yuxuan Shu, Xiao Gu, Guangyao Yang, and Benny P.~L. Lo.
\newblock Revisiting self-supervised contrastive learning for facial expression recognition.
\newblock In \emph{British Machine Vision Conference}, 2022.

\bibitem[Siriwardhana et~al.(2020)Siriwardhana, Kaluarachchi, Billinghurst, and Nanayakkara]{9206016}
Shamane Siriwardhana, Tharindu Kaluarachchi, Mark Billinghurst, and Suranga Nanayakkara.
\newblock Multimodal emotion recognition with transformer-based self supervised feature fusion.
\newblock \emph{IEEE Access}, 8:\penalty0 176274--176285, 2020.

\bibitem[Suess et~al.(2015)Suess, Rabovsky, and Abdel~Rahman]{suessPerceivingEmotionsNeutral2015}
Franziska Suess, Milena Rabovsky, and Rasha Abdel~Rahman.
\newblock Perceiving emotions in neutral faces: Expression processing is biased by affective person knowledge.
\newblock \emph{Social Cognitive and Affective Neuroscience}, 10\penalty0 (4):\penalty0 531--536, 2015.

\bibitem[Sun et~al.(2019)Sun, Myers, Vondrick, Murphy, and Schmid]{Sun2019VideoBERTAJ}
Chen Sun, Austin Myers, Carl Vondrick, Kevin~P. Murphy, and Cordelia Schmid.
\newblock Videobert: A joint model for video and language representation learning.
\newblock \emph{2019 IEEE/CVF International Conference on Computer Vision (ICCV)}, pages 7463--7472, 2019.

\bibitem[Taleb et~al.(2022)Taleb, Kirchler, Monti, and Lippert]{Taleb_2022_CVPR}
Aiham Taleb, Matthias Kirchler, Remo Monti, and Christoph Lippert.
\newblock Contig: Self-supervised multimodal contrastive learning for medical imaging with genetics.
\newblock In \emph{Proceedings of the IEEE/CVF Conference on Computer Vision and Pattern Recognition (CVPR)}, pages 20908--20921, 2022.

\bibitem[Tian et~al.(2020)Tian, Krishnan, and Isola]{cmc}
Yonglong Tian, Dilip Krishnan, and Phillip Isola.
\newblock Contrastive multiview coding.
\newblock In \emph{Computer Vision -- ECCV 2020}, pages 776--794, Cham, 2020. Springer International Publishing.

\bibitem[Wen et~al.(2021)Wen, You, and Fu]{WEN2021103178}
Huanglu Wen, Shaodi You, and Ying Fu.
\newblock Cross-modal dynamic convolution for multi-modal emotion recognition.
\newblock \emph{Journal of Visual Communication and Image Representation}, 78:\penalty0 103178, 2021.

\bibitem[Wieser and Brosch(2012)]{wieserFacesContextReview2012}
Matthias~J. Wieser and Tobias Brosch.
\newblock Faces in {{Context}}: {{A Review}} and {{Systematization}} of {{Contextual Influences}} on {{Affective Face Processing}}.
\newblock \emph{Frontiers in Psychology}, 3, 2012.

\bibitem[Zhang et~al.(2020)Zhang, Tseng, and Kreiman]{zhangPuttingVisualObject2020}
Mengmi Zhang, Claire Tseng, and Gabriel Kreiman.
\newblock Putting {{Visual Object Recognition}} in {{Context}}.
\newblock In \emph{2020 {{IEEE}}/{{CVF Conference}} on {{Computer Vision}} and {{Pattern Recognition}} ({{CVPR}})}, pages 12982--12991, Seattle, WA, USA, 2020. IEEE.

\bibitem[Zheng et~al.(2021)Zheng, Wang, You, Qian, Zhang, Wang, and Xu]{classcollision}
Mingkai Zheng, Fei Wang, Shan You, Chen Qian, Changshui Zhang, Xiaogang Wang, and Chang Xu.
\newblock Weakly supervised contrastive learning.
\newblock In \emph{2021 IEEE/CVF International Conference on Computer Vision (ICCV)}, pages 10022--10031, 2021.

\end{thebibliography}
}

\clearpage
\setcounter{page}{1}
\maketitlesupplementary



\section{Implementation Details}

We extract features from the input modalities using pretrained fixed feature extractors. The frames for the 2D ResNet-152 \cite{He2015DeepRL} (ImageNet \cite{dengImageNetLargescaleHierarchical2009}) are subsampled to 1fps and for the 3D ResNet-101 \cite{pretorchedx} (Kinetics \cite{kinetics}) to 16fps. The audio is transformed into Mel spectrograms before processing with a DAVENet \cite{Harwath2018JointlyDV} pretrained on affective audio. For text, we take the last hidden layer of a DistillBERT \cite{Sanh2019DistilBERTAD} that was trained for sentiment analysis. To obtain a fixed-size feature vector per modality, we process the inputs sequentially and average the resulting vectors over time. The 2D and 3D frame features are concatenated after averaging. The result is stored on disk, \ie no finetuning of the backbone feature extractors happens. We use AdamW optimizer \cite{loshchilovDecoupledWeightDecay2017} for both pertaining and downstream training, in combination with cosine annealing with warm restarts \cite{loshchilovSGDRSTOCHASTICGRADIENT2017} as learning rate scheduling. Our batch size is 4096 and the image size is 180 by 180 pixels (cropped or resized). The size of the representations is 4096. More details are given in the supplementary material. We used the following learning rates in our experiments:

\noindent
\textbf{Pretraining:}

\begin{itemize}
    \item ConCluGen: lr=0.00009, weight decay=0.00032
    \item ConClu: lr=0.00009, weight decay=0.00032
    \item ConGen: lr=0.00036, weight decay=0.00032
    \item Multi-Cont: lr=0.00036, weight decay=
    \item Instance-Cont: lr=0.00036, weight decay=00032
    \item Generative: lr=0.00036, weight decay=0.00032
\end{itemize}

\noindent
\textbf{Downstream:} (We chose common hyperparameters that worked well for all methods)

\begin{itemize}
    \item CAER: lr=0.0061, weight decay=0.08216
    \item MELD: lr=0.00967, weight decay=0.00004
    \item MOSEI: lr=0.00996, weight decay=0.00007
\end{itemize}

For the K-means clustering, we choose a queue size of 4 (\ie 4 batches were considered in the clustering), 8 clusters and started the clustering in epoch 12.

\section{Confusion Matrices for \gls{caer}}

\def\confmatsize{1.2}
\def\tsneconfmatsubfigsize{0.51}
\pgfplotsset{colormap/jet}

\begin{figure}
  \centering
  \begin{subfigure}{0.49\linewidth}
       \includegraphics[width=\textwidth]{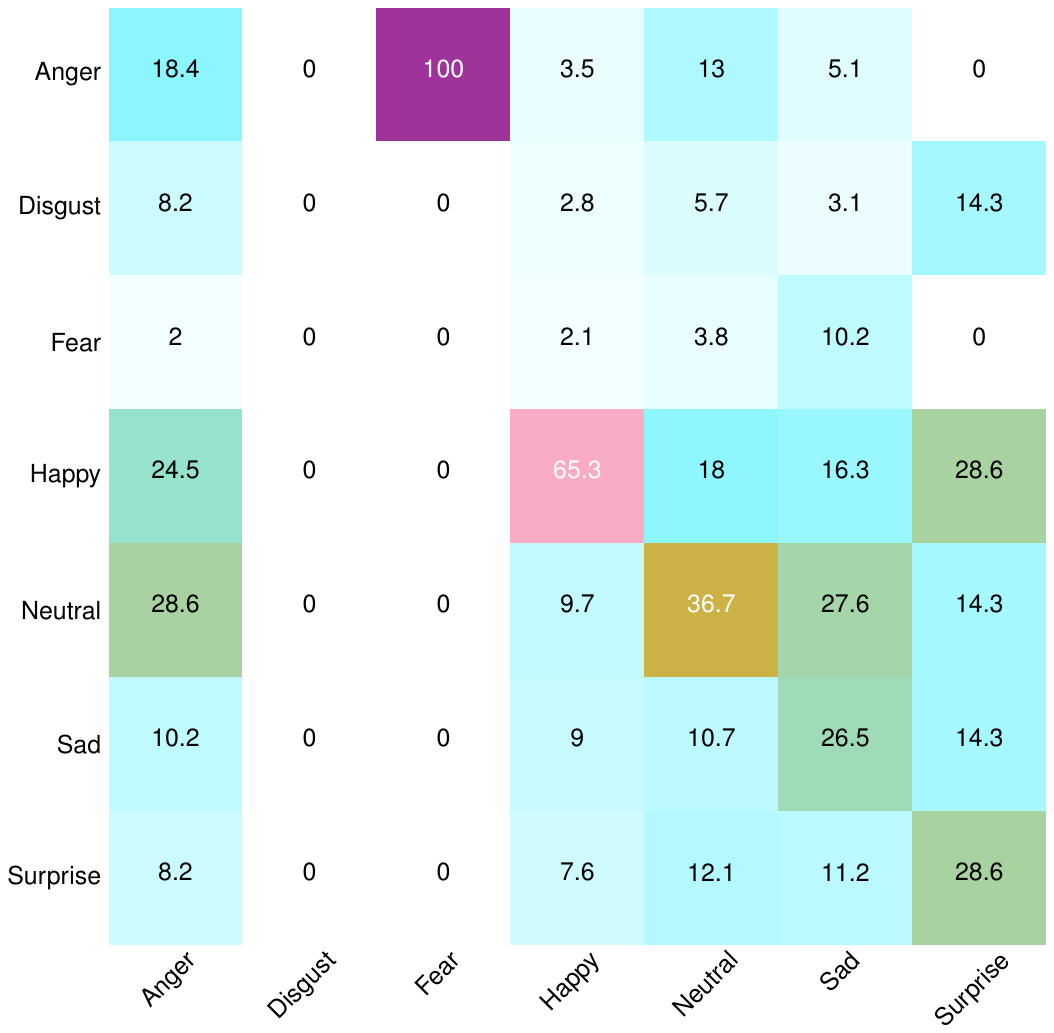}
    \caption{ConCluGen Model.}
    \label{fig:CARE-ConCluGen}
  \end{subfigure}
  \hfill
  \begin{subfigure}{0.49\linewidth}
       \includegraphics[width=\textwidth]{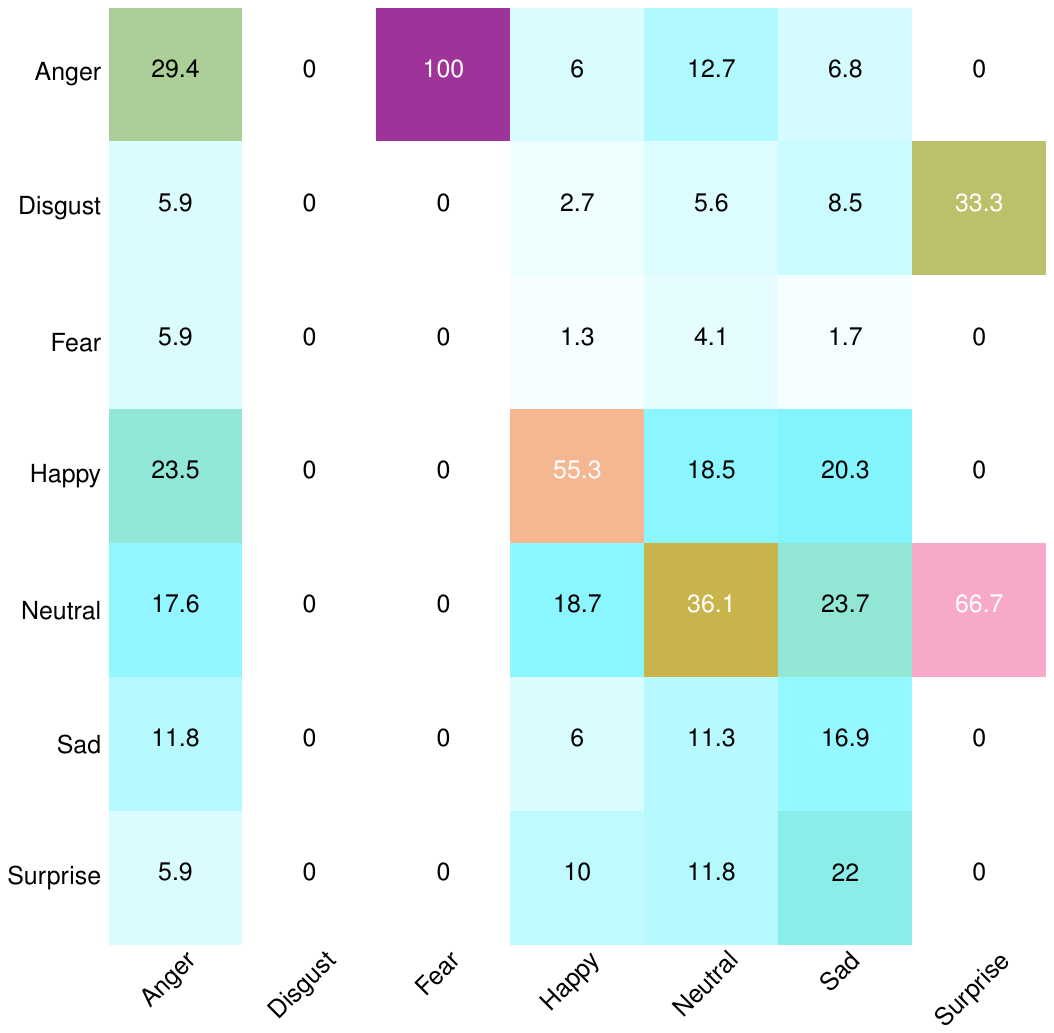}
        \caption{ConClu Model.}
        \label{fig:CARE-ConClu}
    \end{subfigure}
    \hfill
    \begin{subfigure}{0.49\linewidth}
        \includegraphics[width=\textwidth]{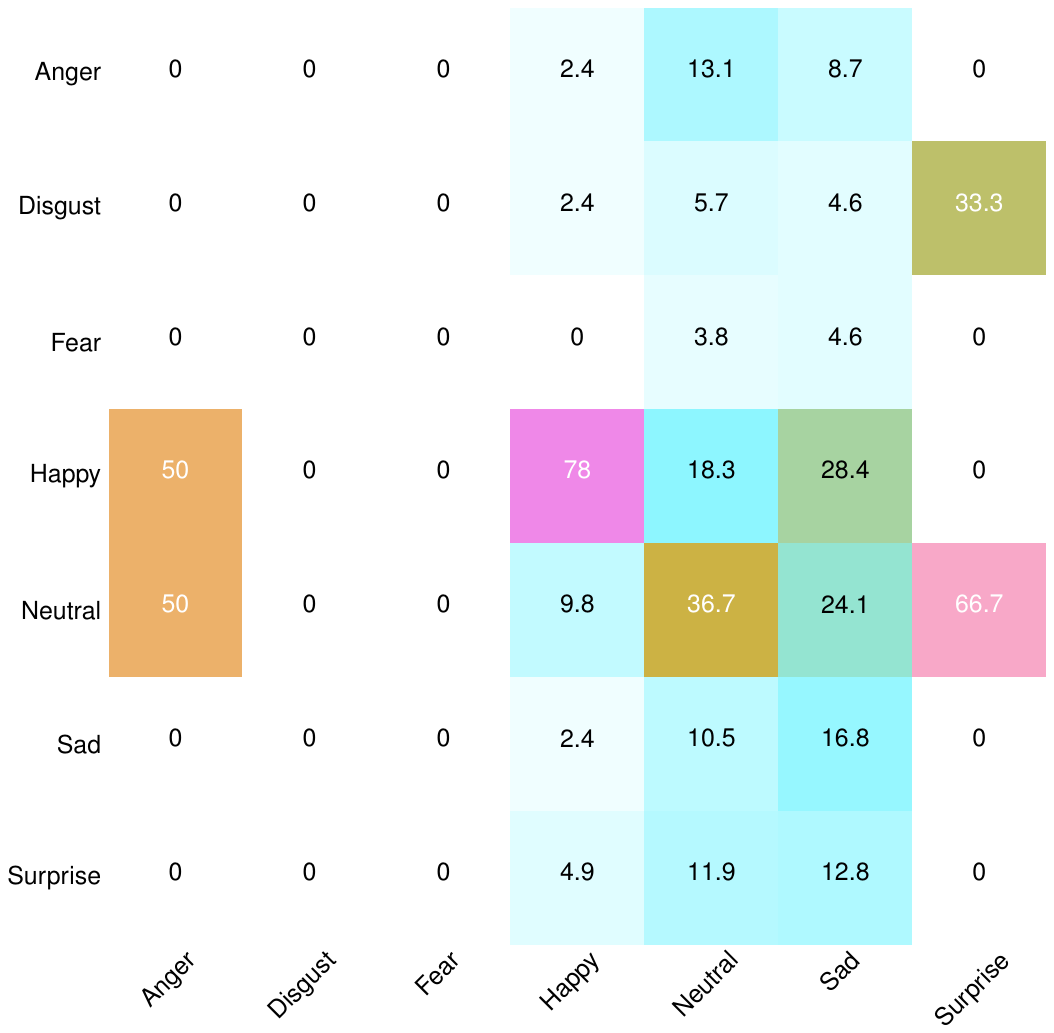}
    \caption{Multi-Cont Model.}
    \label{fig:CARE-Multi-Cont}
    \end{subfigure}
    \hfill
    \begin{subfigure}{0.49\linewidth}
        \includegraphics[width=\textwidth]{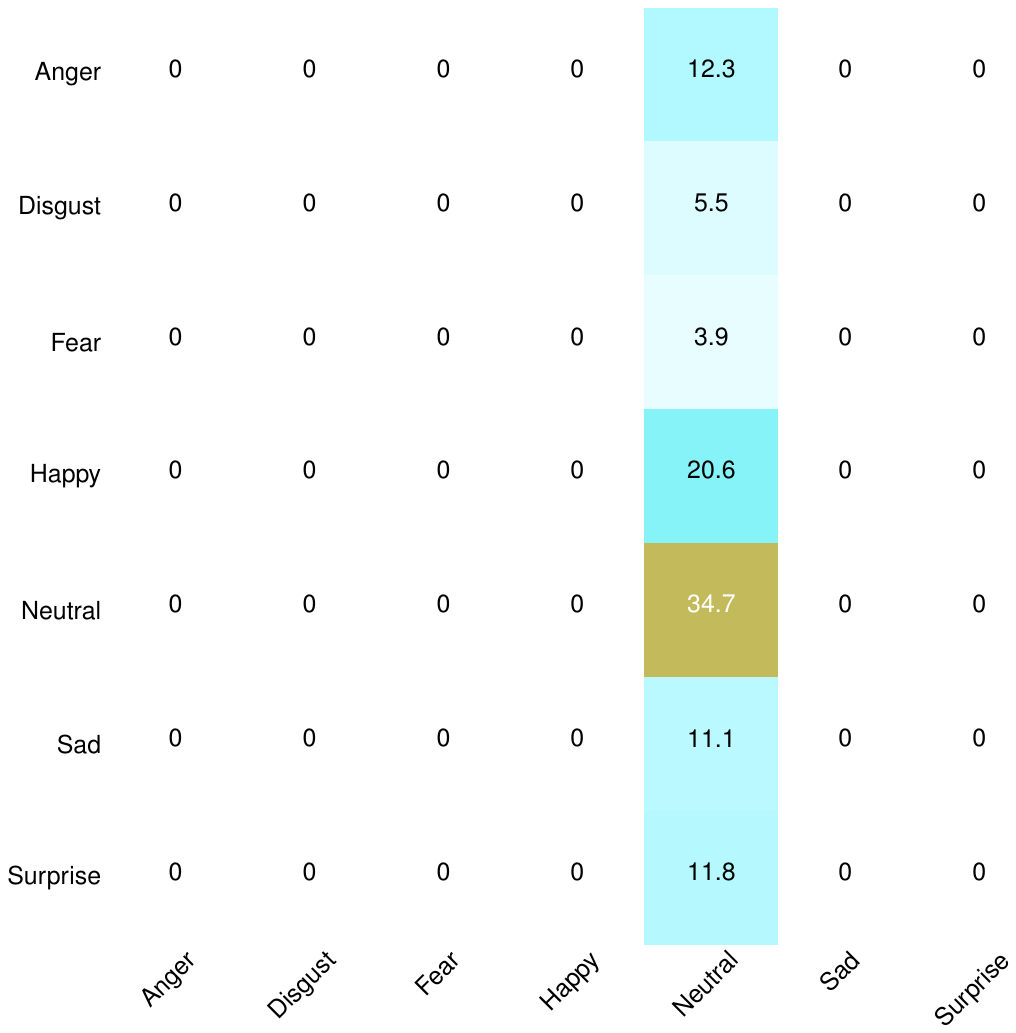}
    \caption{Instance-Cont Model.}
    \label{fig:CARE-Instance-Cont}
  \end{subfigure} 
  \caption{Confusion metrics for CAER dataset over different models.}
  \label{fig:CARE_matrix}
\end{figure}

\end{document}